\documentclass[]{article}

\usepackage{arxiv}

\usepackage[utf8]{inputenc} 
\usepackage[T1]{fontenc}    
\usepackage{hyperref}       
\usepackage{url}            
\usepackage{booktabs}       
\usepackage{amsfonts}       
\usepackage{amsmath}
\usepackage{nicefrac}       
\usepackage{microtype}      

\usepackage{lipsum}         
\usepackage{graphicx}
\usepackage{natbib}
\usepackage{doi}

\usepackage{array}
\usepackage{ragged2e}
\usepackage{siunitx} 
\usepackage{caption} 
\usepackage{comment} 
\usepackage[utf8]{inputenc} 
\usepackage{subfig} 
\newcommand{\tak}[1]{{\textcolor{blue}{ (Tak:~#1})}}

\newcommand{\chase}[1]{{\textcolor{red}{ (Chase:~#1})}}
\title{Harnessing Rich Multi-Modal Data for Spatial-Temporal Homophily-Embedded Graph Learning Across Domains and Localities}

\newif\ifuniqueAffiliation
\uniqueAffiliationtrue

\ifuniqueAffiliation 
\author{ \href{https://orcid.org/0000-0001-5669-8565}{\includegraphics[scale=0.06]{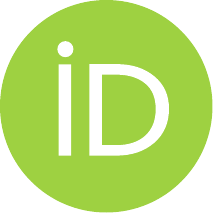}\hspace{1mm}Takuya Kurihana} \\
	Fujitsu Research of America\\
	\And
	Xiaojian Zhang \\
	Department of Civil and Coastal Engineering\\
	University of Florida\\
	\AND
    \href{https://orcid.org/0009-0001-5770-8400}{\includegraphics[scale=0.06]{orcid.pdf}\hspace{1mm}	Wing Yee Au\thanks{This work was presented at the 2025 INFORMS Annual Meeting. Corresponding author. Email: \texttt{wau@fujitsu.com}}}\\
     Fujitsu Research of America\\
	\And
    Hon Yung Wong\\
     Fujitsu Research of America\\
}
\else
\usepackage{authblk}

\newbox{\orcid}\sbox{\orcid}{\includegraphics[scale=0.06]{orcid.pdf}} 
\author[1]{%
	\href{https://orcid.org/0000-0000-0000-0000}{\usebox{\orcid}\hspace{1mm}David S.~Hippocampus\thanks{\texttt{hippo@cs.cranberry-lemon.edu}}}%
}
\author[1,2]{%
	\href{https://orcid.org/0000-0000-0000-0000}{\usebox{\orcid}\hspace{1mm}Elias D.~Striatum\thanks{\texttt{stariate@ee.mount-sheikh.edu}}}%
}
\affil[1]{Department of Computer Science, Cranberry-Lemon University, Pittsburgh, PA 15213}
\affil[2]{Department of Electrical Engineering, Mount-Sheikh University, Santa Narimana, Levand}
\fi

\renewcommand{\shorttitle}{}

\hypersetup{
pdftitle={A template for the arxiv style},
pdfsubject={q-bio.NC, q-bio.QM},
pdfauthor={David S.~Hippocampus, Elias D.~Striatum},
pdfkeywords={First keyword, Second keyword, More},
}

\begin{document}
\maketitle
\begin{abstract}

Modern cities are increasingly reliant on data-driven insights to support decision making in areas such as transportation, public safety and environmental impact. However, city-level data often exists in heterogeneous formats, collected independently by local agencies with diverse objectives and standards. 
Despite their numerous, wide-ranging, and uniformly consumable nature, national-level datasets exhibit significant heterogeneity and multi-modality.
This research proposes a heterogeneous data pipeline that performs cross-domain data fusion over time-varying, spatial-varying and spatial-varying time-series datasets. 
We aim to address complex urban problems across multiple domains and localities by harnessing the rich information over 50 data sources. 
Specifically, our data-learning module integrates homophily from spatial-varying dataset into graph-learning, embedding information of various localities into models. We demonstrate the generalizability and flexibility of the framework through five real-world observations using a variety of publicly accessible datasets (e.g., ride-share, traffic crash, and crime reports) collected from multiple cities.
The results show that our proposed framework demonstrates strong predictive performance while requiring minimal reconfiguration when transferred to new localities or domains. 
This research advances the goal of building data-informed urban systems in a scalable way, addressing one of the most pressing challenges in smart city analytics.

\end{abstract}

\section{INTRODUCTION} 
Urban geospatial data are highly heterogeneous, capturing both the physical and social dimensions of cities across diverse modalities and spatio-temporal resolutions. 
Integrating such multi-modal and complex urban geospatial data supports data-driven urban planning, resource management, and development strategies for relevant stakeholders~\cite{li2024m3}. 
Assembling enriched multi-modal datasets~\cite{chen2024terra, wang2023ssl4eo} may alleviate the significant manual effort involved in data fusion.
However, cross-domain data fusion still involves meticulously handcrafted frameworks, limiting the scalability, generalizability, and adaptation of solutions to diverse downstream domain problems and urban environments with minimal modification.
%

\begin{figure*}[ht!]
    \centering
    \includegraphics[width=.85\linewidth]{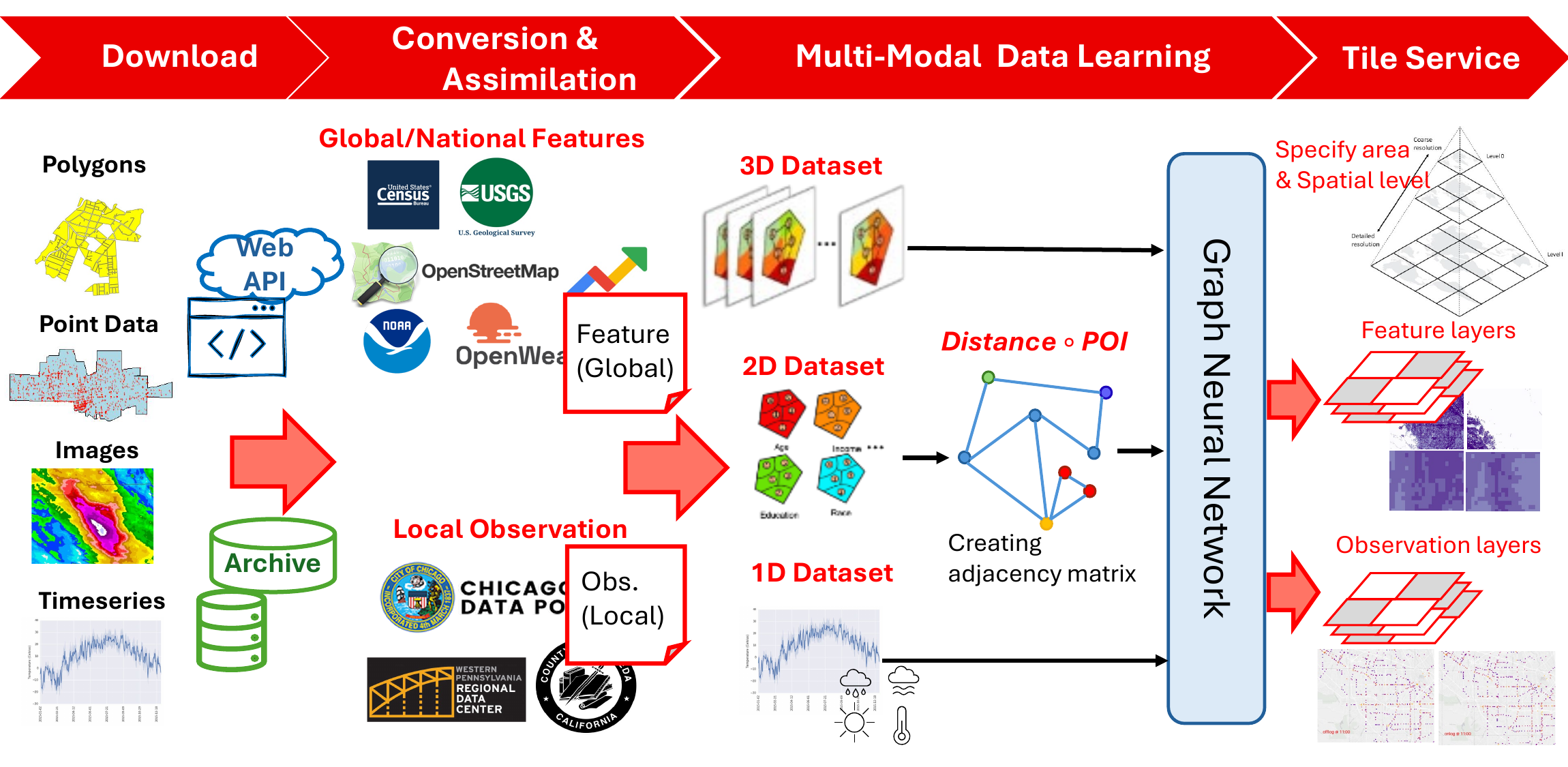}
    \caption{Diagram of our heterogeneous multi-modal graph learning workflow.}
    \label{fig:workflow}
\end{figure*}

Graph-based learning approaches have received increasing attention to represent real-world urban patterns to capture their underlying structure~\cite{hu2024towards, song2020spatial}. 
Graph Convolutional Neural Networks (GCNs) apply graph convolution to extract meaningful spatial patterns from nearby geographical regions when dealing with geospatial data sources such as traffic~\cite{gao2024uncertainty, song2020spatial}, crime~\cite{wang2025uncertainty}, ride-source~\cite{zhuang2022uncertainty}, crash~\cite{zhuang2024sauc} and evacuation ~\cite{zhang2024situational} problems.
Spatio-Temporal Graph Convolutional Networks (STGCNs) incorporate a spatial-temporal module that enables capturing time-series graph features~\cite{yu2017spatio}.
GCNs incorporating zero-inflated frameworks~\cite{wang2025uncertainty,zhuang2022uncertainty} parametrize  distributions with numerous zero data to improve prediction capabilities since urban observation frequently comprise of sparse data.

Real-world urban dynamics has a multifaceted aspect influenced by a wide array of observational factors beyond only spatial proximity~\cite{huang2022unfolding, zou2024learning}.
Recent studies reveal strong
relationships between ride-sourcing demand and a combination of local environment variables, including population density, land use, infrastructure, and transit~\cite{yu2019exploring} as well as socio-economic factors~\cite{lavieri2019modeling} such as income, gender, and ethnicity. 
Hajisafi et al.~\cite{hajisafi2023learning} propose incorporating point-of-interest (POI) correlations within a computational module within GCNs to improve forecasting human behavior and urban dynamics.
These studies suggest the necessity of developing an approach for embedding comprehensive multi-factor urban signals into graph learning frameworks.
However, existing approaches often remain tailored to specific targets, limiting their generalizability \cite{zhang2025towards}.

The homophily of a graph, wherein nodes with similar features tend to be connected by edges, has been shown to enhance the performance of GCNs~\cite{jiang2024self, zhu2020beyond}.
Therefore, the Euclidean distance-based graph construction may not adequately capture local dependencies in urban observation patterns, as it neglects non-Euclidean relationships and socioeconomic factors influencing spatial proximity and similarity among neighborhoods~\cite{wang2022hagen}.
These intra- and inter-region homophily are learnable as embedding representations to capture internal correlations among pairs of nodes through training GCNs~\cite{li2022spatial, liu2020spatiotemporal, wang2022hagen, zhang2022multi}.
Integrating spatial correlation in observation~\cite{zhang2025linear} 
directly incorporates spatial signals inherent in urban systems that may not be fully accounted by localized features via GCNs. 
While such multi-view and heterogeneous graph-based approaches can be effective, they may incur significant computational costs, particularly when transferred to different localities and domains.

For the context in policy and business applications, our goal is aiming at developing a multi-modal data learning framework characterized by multi-spectral capabilities, generalizability, and transferability, enabling its application across diverse geographic locations, problem domains, and data sources.
To accomplish this, we present a heterogeneous multi-modal homophily-embedding graph learning framework to three urban \textit{domains} (i.e. downstream tasks) -- ride-sourcing, crime, and traffic crashes -- and three different \textit{localities} (i.e. cities): Chicago, Pittsburgh, and Oakland.
We develop a workflow that integrates publicly available open datasets into two distinct categories: \textbf{Feature}, a globally available dataset characterized by extensive spatial and temporal coverage, and \textbf{Observation}, diverse local observations with limited spatial and temporal scope archived in municipal government data portals.  
To construct a generalizable multi-contextual graph structure, we propose a homophily embedded graph that leverages correlations among comprehensive socio-demographic and socio-economic variables from extensive number of features.
Our proposed graph creation can adapt a distance-based graph structure into a more contextually appropriate representation.
Our results demonstrate that combinations of multi-modal urban features and observation improve model's predictive capability across three cities.

The rest of the paper is organized as follows. 
Section~\ref{sec:related} summarizes related work for spatio-temporal graph neural networks in urban studies. 
Section~\ref{sec:METHOD} defines the research problem and introduces the proposed data-learning framework. We also describe graph learning methodologies in Section~\ref{sec:METHOD} as well.
Finally, we present our experimental setup and results in Section~\ref{sec:EXPERIMENT}.
\label{sec:intro}

\section{RELATED WORKS}

Cross-domain data fusion, which integrates multi-modal sources of geospatial data into urban studies, is a prominent research topic as the volume of geospatial dataset has dramatically increased with urban developments~\cite{fadhel2024comprehensive}. 
Commonly employed data fusion approaches can be categorized into feature-based, stage-based, and semantic meaning-based data fusion, based on the stage at which data is fused in artificial intelligence (AI) frameworks and the similarities of the input data~\cite{liu2020urban, zheng2015methodologies}.
The advancement of AI techniques open up novel data fusion approaches~\cite{zou2025deep}.
For graph-based learning, recent works propose data fusion approaches on graph. Graph-based data fusion offers multi-view graph network-based data fusion by compounding connectivity among multiple graphs~\cite{zhang2022multi}.
Heterogeneous graph-based data fusion addresses limitations of multi-view approach in struggling to capture high-dimensional and cross-modal correlations due to the lack of direct interaction.
Liu et al.~\cite{liu2020spatiotemporal} introduce hierarchical graph embedding to capture high-order interaction among various data layers. 
ST-SHN~\cite{li2022spatial} designs a graph-structured message passing layer to generate a hypergraph representation for semantic crime relationships.
GeoHG~\cite{zou2024learning} proposes heterogeneous graph-based structure to embed spatial, built-in environmental, societal, socio-economic information into a graph. 
Despite existing research concentrates on specific task domain, it requires further effort to build a holistic approach for diverse urban challenges. 

Graph convolutional neural networks (GCNs) are powerful data-driven methods to address a wide range of complex urban challenges through developing novel computational modules. 
T-GCN~\cite{zhao2019t} employs gated recurrent network (GRU) and graph convolution.  
DCRNN~\cite{li2017diffusion} approximates traffic system as random walks to implement graph diffusion convolution. 
GLDNet~\cite{zhang2020graph} introduces localized diffusion convolution to control spatial propagation of crime events dependent on the heterogeneous network topology. 
Combination of GRU with diffusion convolution captures neighborhoods' dependencies~\cite{sun2021crimeforecaster}. 
STGCN~\cite{yan2018spatial} and STSGCN~\cite{song2020spatial} introduce a Chebyshev polynomial as a graph convolution operation and stack spatial-temporal blocks. 
AIST~\cite{rayhan2023aist} combine crime categories and external features from traffic flow and various POI categories. 
HAGEN~\cite{wang2022hagen} incorporates intra- and inter-regional correlation based on POI and distance feature dynamically into graph learning.
Recent works integrates attention mechanism into spatial and temporal graph convolutional networks to enhance their learning capabilities~\cite{lin2024channel, tang2022short,yang2021tctn}
Zero-inflated model has been applied to the GCN frameworks in various urban problems such as ride-source~\cite{zhuang2022uncertainty}, crime~\cite{wang2025uncertainty}, and traffic~\cite{gao2024uncertainty} to handle sparse observation.

\label{sec:related}

\section{METHOD}\label{sec:METHOD}

Our heterogeneous multi-modal homophily-embedded graph learning framework presented in~\autoref{fig:workflow} is aiming at fusing various multifaceted datasets upon their modalities as well as informing urban built-in environment to graph. 
We therefore define two hierarchal data categories:  \textbf{Feature} that depicts nationally and globally available data and \textbf{Observations} that is only locally (i.e., city and county level) available.
The `Feature' data provides the interaction of cross-domain connections. And `Observation' collected from target domains by local agencies provides spatio-temporal data and their dependencies.  
Subsequently, we transform the Features and Observations into \textit{1-dimensional} (1D: time series data) for informing temporal information shared over a target domain, 2-dimensional (2D: time-constant spatial data) and 3-dimensional (3D: spatiotemoral time series data) into three dataset asset classes to accommodate different modality inputs within graph learning for predicting the state of a target domain (i.e. ride-sourcing, crime, and crash) at a single time step in the future.



\subsection{Feature Data Collection}\label{sec:feature_dataset}
Leveraging global and nation-wide dataset allows data processing pipelines to be transferrable among different domains and localities with minimum adaptation.
To achieve this purpose, we collect demographics, economy, road network, land cover and weather features. 
The feature data categories are organized as follows, along with their corresponding modalities

\textbf{Demographics (2D)}: We select \num{132} features from population, race, age, housing, income, mobility, and employment collected from the 2020 American Community Survey (ACS) and Decennial Census of Population and Housing (DCPH) in the United States Census Bureau (USCB)~\cite{uscb_acs}. 
In this study, we work with `census tract', a unit of geographic resolution for census study. 

\textbf{Economy (2D)}: We resample the employment data from the Longitudinal Employer-Household Dynamics (LEHD)~\cite{uscb_lehd} in USCB. 
USCB defines 20 industry categories and we classify them into five primary industries: \textit{Retail}~(retail trade), \textit{Office} (IT, finance/insurance, real estate, enterprise), \textit{Industry}~(agriculture, forestry, fishing/hunting, mining, oil/gas, utilities, construction, manufacturing, and wholesale trade), \textit{Service}~(professional/scientific/technical/educational services, administrative/support, waste management, and other), and \textit{Entertainment}~(arts, entertainment, accommodation, restaurant). 

\textbf{Road network (2D)}: We calculate road densities and intersection density per square miles from OpenStreetMap Overpass~\cite{OpenStreetMap}.

\textbf{Land cover (2D)}: The National Land Cover Database~\cite{homer2012national} collected from United States Geological Service National Land (USGS) provides 16 land use categories. We simplify them into four categories: \textit{developed}, \textit{cultivated}, \textit{water}~(water and wetlands), and \textit{vegetation}~(barren, forest, shrubland, herbaceous) area. 

\textbf{Weather (1D)}: While some weather conditions (e.g., precipitation) may vary in neighborhoods even within the same city, temporal variations have a greater impact on urban dynamics than spatial factors.
In particular, we select temperature, humidity, wind speed and direction, rain, and snow depth at the surface level. 
We combine the amount of rain and snow due to the sparse observations. 
Hourly point weather data is downloaded from OpenWeatherMap ~\cite{openweather}.

\subsection{Homophily and Distance-based Adjacency Matrix}\label{sec:graph_adj}
\begin{figure}[htbp!]
    \centering
    \includegraphics[clip, width=.85\columnwidth]{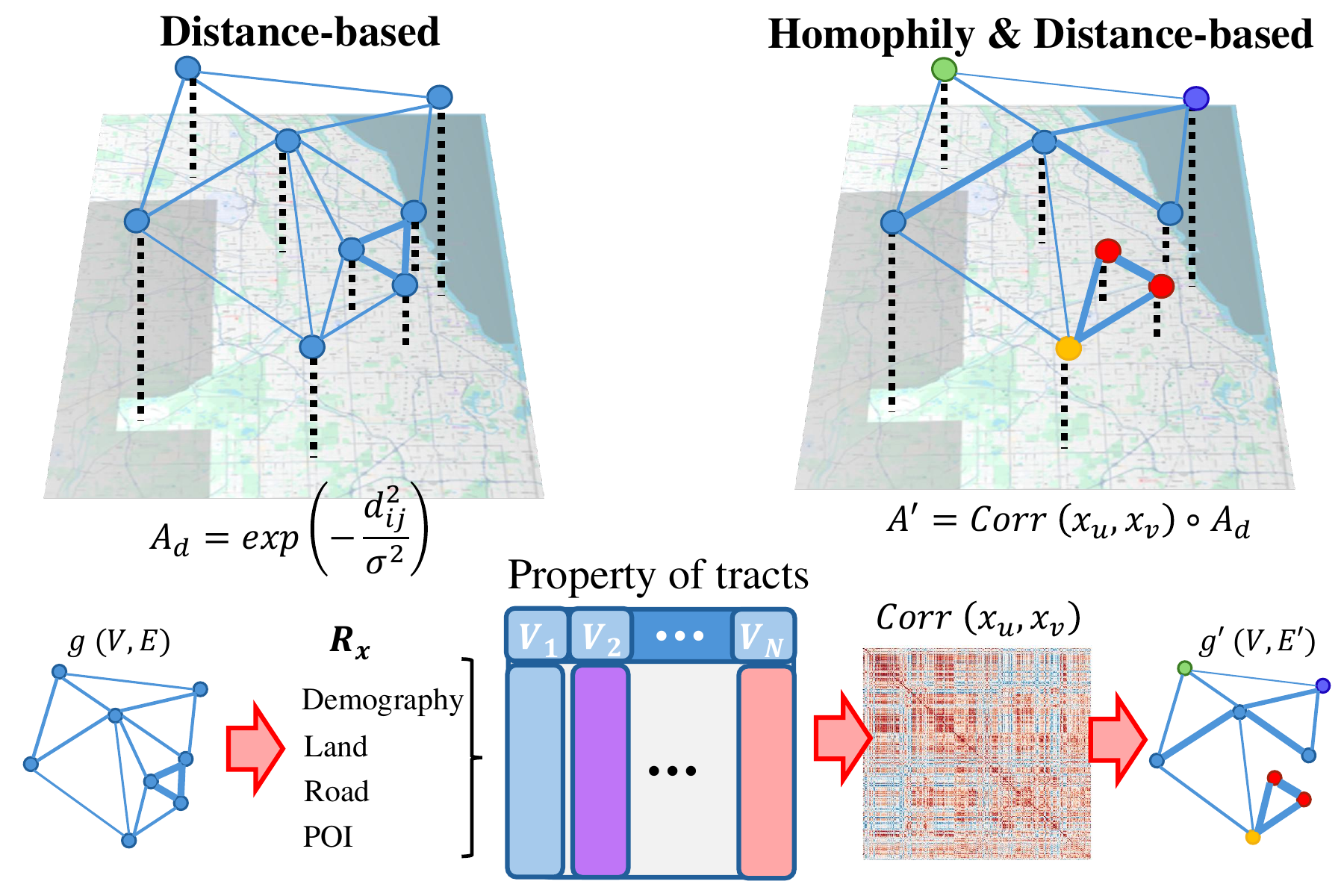}
    \caption{Illustration of the creation of homophily embedded graph structure. Distance-based graph is weighted by correlations derived from 48 socio-economic, socio-demographic, and environmental features among census-tracts, allowing graph networks to reflect multi-contextual information.}
    \label{fig:city-graph}
\end{figure}
Standard approach to create graph is using regions as vertices (hereafter we use `node' and `vertex' interchangeably) and edges are defined with spatial adjacency or some similarity metrics between any pair of two vertices.
The adjacency matrix of the distance-based graph~\cite{yu2017spatio} is defined as,
\begin{eqnarray}\label{eq:graphd}
    A_d =
\begin{cases}
\exp{\left( -\frac{d_{ij}^2}{\sigma^2} \right)} &  i \neq j \text{ and } \exp{\left( -\frac{d_{ij}^2}{\sigma^2} \right)}\geq \epsilon_d \\
0 &,  o.w.
\end{cases}
\end{eqnarray}

where $A_d$ is the adjacency matrix, which is determined by a Euclidean distance of a pair of $i^{\mathrm{th}}$ tract and  $j^{\mathrm{th}}$ tract, $\epsilon_d$ is a threshold in the range of $[0,1]$, and a control variable $\sigma$.  
We assign $\epsilon_d = 0.3$ and $\sigma = 10$ from our parameter search, respectively.

However, cities present complex structures, thereby graph created only from distance-based approach is not sufficient to represent the characteristic of urban areas. 
For example, it is often observed that two neighboring regions have different socio-economic patterns~\cite{wang2022hagen}.
One solution is to utilize heterogeneous graph-based structure~\cite{zhang2022multi, zou2024learning} as a part of graph learning module, while effectively learning special interactions through graph embeddings requires a sufficient number of data to ensure meaningful results.
One potential solution is multi-graph convolutions, where each graph is constructed on pre-calculated correlations derived from individual properties~\cite{zhang2025linear}, whereas the method may exhibit the scalability limitation as increasing the number of correlation graphs when attempting to model high-dimensional urban statistics. 

Therefore, we introduce a more transferable solution, \textbf{homophily-embedded graph}, that adapts the weights of a distance-based adjacency matrix $A_d$ using a correlation matrix that incorporates an extensive number of explainable variables in a city.
To achieve this goal, we create a Pearson correlation coefficient matrix $\mathrm{Corr} \in \mathcal{R}^{N \times N}$ between combinations of two nodes $V_i$ and $V_i$ among $N = |V|$ nodes based on derived 48 features spanning from demographics, land cover, and POI (economy  and road network), respectively (See~\autoref{sec:feature_dataset} in details). 
We derive custom features by aggregating a selection of the original features from data sources.
Note that walk score feature is only included for case studies in Chicago.   
Subsequently, we aggregate and normalize the matrices; 
\begin{equation}
    A' =  \frac{1}{|S|}\sum_{C \in S } \mathrm{Corr (X_{V_i}, X_{V_j})} \circ A_d ,
\end{equation}
where $S$ is a set of correlation graphs (i.e., demography, land cover, and POI);
$\mathrm{Corr} = \frac{\mathrm{Cov}_c(X_{V_i}, X_{V_j})}{ \sigma_{X_{V_i}} \cdot \sigma_{X_{V_j}} } $ is a correlation matrix; $\mathrm{Cov}$ is covariance, $\sigma$ depicts a standard deviation of features in a node; $A_d$ is a distance-based adjacency matrix from~\autoref{eq:graphd}; and $X_{V_i}$ and $X_{V_j}$ represent a set of 48 features listed in~\autoref{tab:variable_names} (See~\autoref{tab:full_variable_names} in details) from node $ (V_i,V_j)\in \mathcal{G}(V, E)$; and $\circ$ represents the Hadamard matrix operation.
Note that we consider only the magnitude of the correlation coefficient (i.e., ignore the sign of correlation).

\begin{table}[htbp!]
\centering
\footnotesize
\caption{Summary of features calculating correlations among census tracts. See~\autoref{tab:full_variable_names} for a comprehensive list of all variables.} 
\begin{tabular}{p{1cm} | p{2.2cm}p{1.2cm}p{5cm}p{1.5cm}}
\toprule
Category &  Title & \# Feature & Description & Data source  \\
\midrule
Demo. & Population & 10 & Total population, density, and key demographic variables (sex, race, and age) & USCB \\
      & Household & 8 & Household size, family type, income, \# owned cars, house/rent value & \\
      & Education & 3 & Percentage of level of degrees & \\
      & Commuting & 3 & Commute time, type of commuting, \# transit & \\
      & Employment & 10 & Employment status and income level &  \\
      & Wealth & 1 & Gini-index & \\
      & Housing & 1 & Renter-occupied housing units & \\ \midrule
 Econ. & Occupation & 5 & Retail, Office, Service, Entertain, and other Industry & LEHD/USCB \\ \midrule
 Road & Road network & 3 & Density of roads and intersections, walkability & OpenStreetMap,  Walkscore.com \\ \midrule
 Land & Land use& 4 & Area of water, developed, cultivated, and vegetation area & USGS \\ 
\bottomrule
\end{tabular}
\parbox[t]{0.98\columnwidth}
{\vskip3pt{\footnotesize $^*$Note: LEHD: Longitudinal Employer-Household Dynamics; USGS: United States Geological Survey. USCB: United States Census Bureau}}
\label{tab:variable_names}
\end{table}

\begin{figure}[htbp!]
    \centering
    \includegraphics[clip, width=.95\columnwidth]{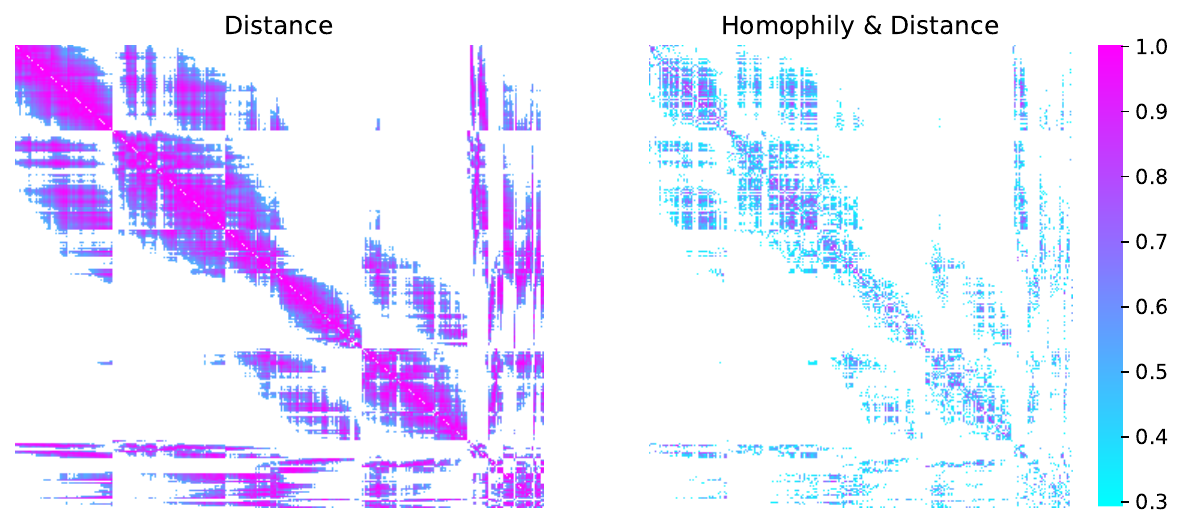}
    \caption{Heatmaps of the weights of adjacency matrix resulting from distance-based (\textbf{left}) and homophily-embedding (\textbf{right}) approaches.}
    \label{fig:adj_weight}
\end{figure}

\autoref{fig:adj_weight} displays an example of structure of adjacency matrices with and without homophily-embedding in Chicago.
The figure indicate that weighting by correlations truncate spurious  or remain relevant relationships among nodes. 
See~\autoref{append:homo} for the weights of adjacency matrices in the other cities.

\subsection{STGCNs for Prediction}\label{sec:stgcn_arch}
For spatio-temporal graph learning, our data-learning module~\autoref{fig:workflow} employs  Spatio-Temporal Graph Convolution Network (STGCN)~\cite{yan2018spatial}, which consists of a sequence of temporal and spatial convolutional block repeatedly and a fully connected output layer to predict next future timestep. 

The ST-block first implements a temporal convolutional layer using the temporal gated convolution.
The input of temporal convolution is $T$ length sequence with $C^{l-1}$ channels such that $h^{(l-1)} \in \mathbb{R}^{T \times N \times C^{l-1}}$ of block $l-1$. 
The output of temporal block can be defined as, 
\begin{equation}
    h^{(l-1)}_{tmp}  
    =  (W_f h^{(l-1)} + b_f) \odot \mathrm{sigmoid}(W_g h^{(l-1)} + b_g), 
\end{equation}
where $W_f \in \mathbb{R}^{C\times C_o}$ and $W_g \in \mathbb{R}^{C\times C_o}$ are learnable parameters ($C_o$ is a size of output channel); $b_f \in 
 \mathbb{R}^{C_o}$ and $b_g\in 
 \mathbb{R}^{C_o}$ are biases for filter ($f$) an gate ($g$), respectively. $\mathrm{sigmoid}$ denotes sigmoid function $\frac{1}{1+e^{-x}}$. 

The subsequent spatial block for the output $h^{l} \in \mathbb{R}^{(T - 2(k - 1)) \times N \times C^{l}}$ uses graph convolution operation defined as,
\begin{equation} 
 h^{(l)} = \mathrm{ReLU}(\Theta*_{\mathcal{G}} h^{(l-1)}_{tmp} ),
\end{equation}
where  $\Theta*_{\mathcal{G}}$ is a spectral kernel of graph convolution with a kernel size $k =3$.  
Details regarding the graph convolution operation can be found in Yu et al.~\cite{yu2017spatio}.
To incorporate 1D features and 3D observation, we employ feature concatenation at the input layer to fuse these two modalities.  

We use Huber loss to optimize the weights of STGCN. 
The Huber loss function \( L_{\delta} \) is defined as:
\begin{equation}
L_{\delta}(y,\hat{y} ) = 
\begin{cases} 
\frac{1}{2}(y - \hat{y} )^2 & \text{for } |y - \hat{y}| \leq \delta \\
\delta |y - \hat{y}| - \frac{1}{2} \delta^2 & \text{for } |y - \hat{y}| > \delta
\end{cases}    
\end{equation}
where $\hat{y}$ denotes an output from STGCN. $\delta_h = 10$ is the threshold parameter after conducting hyperparameter search.

\subsection{STZINBs for Prediction}\label{sec:stzinb_arch}

Zero-inflated model is introduced to handle sparse observation (i.e. the majority of observation is zero). 
It is commonly assumed that crash and crimes follow the zero-inflated negative binominal (ZINB) distribution~\cite{wang2025uncertainty, gao2024uncertainty} defined as,
\begin{equation}\label{eq:zinb_dist}
\begin{aligned}
    f_{\mathrm{ZINB}}(y:\pi, n, p)  & =& \\
    P(Y = y) & = &
\begin{cases} 
\pi + (1 - \pi)(1 - p)^n & \text{if } y = 0 \\
(1 - \pi) \binom{y + n - 1}{y} p^y (1 - p)^n & \text{if } y >0 
\end{cases}
\end{aligned}
\end{equation}
where $\pi$ represents the probability of zeros (zero inflation), reflecting the likelihood of observing no occurrences of the target variable. 
The parameters $r$ and $p$ determine the shape parameters of the traditional negative binomial distribution. 
We use the spatial-temporal zero-inflated negative binominal GCN (STZINB)~\cite{zhuang2022uncertainty} to predict next future timestep of a target observation. 
The STZINB estimates the three parameterized $n$, $p$, $\pi$ from spatial and temporal blocks, respectively.
Finally, the Hadamard product is applied to $n$, $p$, and $\pi$ from the respective spatial and temporal blocks to generate the distribution parameters.

The spatial block of STGCN is composed of Diffusion Graph Convolution Network (DGCN)~\cite{li2017diffusion}. 
The output of GCN at the block $l$ is defined as,
\begin{equation}
    h^{(l+1)} = \sigma\left( \sum_{k=1}^K T_k (\tilde{W_f})h^{(l)}\Theta^k_{f,l} + T_k(\tilde{W_b})h^{(l)}\Theta^k_{b,l} )\right),
\end{equation}
where $T_k(\cdot)$ denotes the Chebyshev polynomial that approximates the convolution operation in DGCN;
$\tilde{W_f} = A'/\mathrm{rowsum(A')}$ depicts the forward transition matrix in forward diffusion, and the backward transition matrix in backward diffusion is $\tilde{W_b} = \tilde{W_f}$ as our adjacency matrix $A$ is symmetric.  
$\Theta$ is learned parameters that control how each node transforms information in forward ($f$) and backward ($b$); $\sigma$ denotes an activation function. We use ReLU in this study. 

The temporal block of STZINB uses Temporal Convolutional Network (TCN)~\cite{lea2017temporal} that applies a gated 1D convolution with width $w_l$ in block $l$.
TCN layer is defined as,
\begin{equation}
    h^(l) = \sigma\left( \Gamma_l * h^{(l-1)} + b\right),
\end{equation}
where $\Gamma_l$ is the 1D convolution filter with width $w_l$ in the $l^{\mathrm{th}}$ layer; $b$ represents a bias. 
TCN layer applies a gated 1D convolution with width $w_l$. Thus, in the $l^{\mathrm{th}}$ layer $h_l \in \mathbb{R}^{B\times N \times w_{l}} $ and $\Gamma \in \mathbb{R}^{w_l \times w_{l-1}} $. Following a community standard, we stuck 3 layers of DCGN and TCN layers, respectively.

To optimize the parameterized zero-inflated distribution, the network leverages variational inference to maximize the log-likelihood, as follows,
\begin{equation}\label{eq:likelihood}
\mathcal{L}_y = 
\begin{cases}
\log \pi + \log(1 - \pi) p^n  & \text{when } y = 0 \\
\log(1 - \pi) + \log \Gamma(n + y) - \log \Gamma(y + 1) & \\
 \text{ }\text{ } - \log \Gamma(n) + n \log p + y \log(1 - \pi) & \text{when } y > 0
\end{cases}
\end{equation}
where $\Gamma$ here denotes the Gamma function, the loss function $\mathcal{L}_y$ is conditionally calculated for $n$, $p$, and $pi$ according to the index of $y=0$ and $y>0$, respectively.
The final loss form is written as,
\begin{equation}
    \mathcal{L} = - \mathcal{L}_{y=0} - \mathcal{L}_{y>0} .
\end{equation}

\subsection{STZINB with Attention layer}\label{sec:stzinb_arch2}

Adapting attention-module to geospatial problems has been widely studied to capture complex dependencies of spatio-temporal relationships~\cite{liu2023spatio}. 
To better incorporate 1D weather features into STZINB model,
we adapt a cross-attention module and apply it before calculating the Hadamard product of the final layer's output.
Particularly, we apply a multi-head cross-attention layer to embed the time series weather trends into parameter embeddings $n$, $p$, $\pi$ from STZINB, respectively.
The attention computation is written as 
\begin{equation}
\begin{aligned}
    Q = X_{1d}&W_Q,K = h_l W_K, V = h_l V_K, \\
    \mathrm{Attn} &= \mathrm{softmax}\left(\frac{QK^T}{\sqrt{d_k/h}}\right) V,     
\end{aligned}
\end{equation}
where $X_1d$ is 1D input data; $W_Q, W_K, W_V \in \mathbb{R}^{d_h \times (d_h/h)}$, $K$ are learnable parameters; $d_h$ is the embedding demission; $h$ is number of attention heads.
A feed-forward layer follows the cross attention computation.
Due to a limited training sample, we do not repeatedly combine the transformer layers to remain the network parameters small.
Note that we also examined through the hyperparameter search to determine the number of transformer layers as 1.

\label{sec:method}

\section{EXPERIMENTS}\label{sec:EXPERIMENT}
In this section, we elaborate data collection for observation,  experiment setups, and results of our multi-modal graph learning framework.
\begin{table}[h!]
\centering
\small
\caption{Summary of datasets}
\begin{tabular}{p{2.5cm}|p{.9cm}p{4.4cm}p{.9cm}p{.85cm}}
\toprule
Dataset  &  Window & Data length & \#Tracts &  Zero rate ($\%$) \\
\midrule
 CDP\_Ride\_1h  & 1 hour & 2018/11/1 --2019/3/31 & 771 & 25 \\
 CDP\_Crime\_4h & 4 hour & " & " & 88 \\
CDP\_Crash\_4h  & 4 hour & " & " & 94 \\
WPRDC\_Crime\_1D & 1 day & 2016/8/1--2023/11/14 & 236 & 97 \\
ACODH\_Crime\_6h & 6 hour & 2012/1/1 --2022/4/2 &340 & 93 \\
\bottomrule
\end{tabular}
\label{tab:dataset}
\end{table}

\subsection{Observation Data Collection}\label{sec:data}

This study focuses on three United States cities (Chicago, Oakland, and Pittsburgh) representing a range of population sizes and localities. 
To assess the generalizability of our framework, we select ride-sourcing, crime, and traffic crash observations in Chicago. 
To further assess its transferability, we evaluate crime occurrence prediction across Chicago, Oakland, and Pittsburgh. 
The five case study datasets are collected from the City of Chicago Data Portal (CDP)\footnote{https://data.cityofchicago.org/}, Western PA Regional Data Center (WPRDC)\footnote{https://www.wprdc.org/en}, and the Alameda County Open Data Hub (ACODH)\footnote{https://data.acgov.org/}. 

Transportation Network Providers (\textbf{CDP Ride}) dataset provides the trip records, including pickup and drop census tract locations and timestamps reported every 15 minutes by ride-sourcing companies.
The CDP Ride dataset does not contain any information that directly identifies the passenger characteristics. 
We select a 5-month sample from November 1st, 2018 to March 31st, 2019. 
We further aggregate the observation at an hourly temporal resolution based on the number of pickups at each tract. 
The sample includes 771 tracts within 1331 tracts in Cook county.
While conventional ride-sourcing studies sample O-D pairs ~\cite{gao2024uncertainty, zhang2022machine}, our study focuses on the number of demands per tract for prediction task, which is known as the trip generation problem in travel demand modeling tasks \cite{zhang2024travel, ben1996travel}.  Traffic crashes (\textbf{CDP Crash}) dataset provides individual traffic crash report from the Chicago Police Department.
The reported crashes are limited cases with a property damage value \$\num{1500} or more, or involving injury to any person by moving vehicles. Crimes (\textbf{CDP Crime}) dataset reflects crime incidents and their block level locations from 2001 to present in Chicago reported by Chicago Police Department. 
For the privacy protection purpose, reported longitude and latitudes are slightly 
shifted from the exact location but kept within the same block.
The sample areas and time-length of CDP crash and CDP crime datasets align with one used for the CDP ride to reuse the same weather and adjacency matrix in modeling.
For aligning data spatially, we identify the census tract that spatially corresponds to the reported latitude and longitude coordinates.
However, compared to the CDP ride data, the density of crime and crash occurrences are less frequent,  so that we aggregate the dataset by 4-hour time-interval. 

The Pittsburgh Police Arrest Data (\textbf{WPRDC Crime}) dataset contains arrest information ranging from serious felony to a failure to appear for trial by block level reported by the City of Pittsburgh Police from March 11, 1998 to present. 
We use the WPRDC crime data from August 1st, 2016 to November 14th, 2023 by aggregating reports at daily cadence per census tract.

The Alameda County Sheriff's Office crime reports (\textbf{ACODH Crime}) provides crime reports at block level longitude and latitude 
covering the period January 1st, 2012 to present. 
We select the ACODH crime data from January 1st, 2012 to April 2nd, 2022 by aggregating reports for 6-hour time-interval at tract level. 

\autoref{tab:dataset} summarizes the time-window, the period of data, the number of tracts (i.e., nodes) and the number of zero rate.    
We set the time-interval to balance by reducing the zero rate but contain diurnal cycle of events that could influence of patterns in crime and crashes. 
CDP ride data only contains 25\% of zero, while the other crash and crime datasets show 88\% to 97\% of zero rate even after the resampling process.  
Due to the lower amount of reports, we apply the daily interval to WPRDC crime dataset.

\subsection{Evaluation Metrics}\label{sec:metrics}
To evaluate the model performance for prediction tasks, we use the Mean Absolute Error (MAE):
\begin{equation}
    \mathrm{MAE} = \frac{1}{|V|} \sum^{|V|}_{i=1} | y_i - \hat{y}_i | ,
\end{equation}
where $y_i$ and $\hat{y}_i$ are the ground-truth and predicted values of $i^{\mathrm{th}}$ data point at a timestep. When evaluating MAE, we denormalize the predicted values to compare in original units. A smaller MAE represents better performance. 
Given the nature of city population density where only few tracts and regions have the order of magnitude of more population, we calculate MAE over the all census tracts \textit{MAE(tract)} and five most populated tracts \textit{MAE(downtown)} for performance comparison (hereafter we refer to these highly populated areas as `downtown' as a convenience).  

To account for percentage error difference in spatial structure since error always high in only few populated area, we calculate the Mean Absolute Percentage Error per tract but due to 0 values in observation, we handle where the actual value is 0 by dividing by 1 instead for a workaround to avoid division by zero,
\begin{equation}\label{eq:mape}
    \mathrm{MAPE} = \frac{1}{|T|} \sum_{i=1}^{|T|} \left| \frac{y_i - \hat{y}_i}{\max(y_i, 1)} \right| \cdot 100, 
\end{equation}
where $T$ is the number of total time steps. Thus, we compute the percentage error over each tract to analyze the spatial structures with and without homophily-embedded adjacency matrix as well as 1D weather features. 

We use the Kullback-Leibler Divergence (KL-Divergence) to evaluate the similarity of distribution between model outputs and inputs besides MAE. 
KL-Divergence is defined as:
\begin{equation}
    \mathrm{KL-Div.} = \frac{1}{|V|} \sum_{i=1}^{|V|} y \log{ \frac{y+ \epsilon}{\hat{y} + \epsilon}}, 
\end{equation}
where a constant $\epsilon = 2.2e-16$ avoids the zero-division since KL-divergence needs to be calculated for non-zero entries. A smaller KL-divergence means the predicted and ground-truth data show similar distributions. 

For evaluating uncertainty with the zero-inflated model, we calculate the Mean Prediction Interval Width (MPIW) on the 10\%--90\% confidence interval:
\begin{equation}
    \mathrm{MPIW}  = \frac{1}{|V|} \sum_{i=1}^{|V|} (U_i - L_i) ,     
\end{equation}
where $U_i$ and $L_i$ are the lower and upper bound of confidence interval of $i^{\mathrm{th}}$ data point respectively. 

Alongside with MPIW, we use Prediction Interval Coverage Probability (PICP) to assess the reliability of prediction intervals if models provide a realistic range of predictions within the given intervals. Specifically, We calculate the PICP with 10\%--90\% confidence interval. 
\begin{equation}
\mathrm{PICP} = \frac{1}{N} \sum_{i=1}^{N} \mathbb{I}(\hat{y}_i \in [y_i^{L}, y_i^{U}]), 
\end{equation}
where $y_i^{L}$ and $y_i^{U}$ represent the lower and upper bound of ground-truth observation. The metrics is optimal when coverage of model output is as close to 80\% as possible.

Since observation in our case studies provide discrete values, we use the F1-score to assess the accuracy of prediction.
In particular for the sparsity of crash and crime observation shown in~\autoref{tab:dataset}, we calculate the true-zero rate to quantify the accuracy of model prediction for zero event.
Model outputs are rounded to the nearest integer in this metrics when calculating F1-score and true-zero rate.
Larger F1-score and true-zero mean more accurate model performance.

\subsection{Experiment Setups}
For evaluating model performances, we compare the metrics defined in~\autoref{sec:metrics} for prediction values and the ground-truth at next future timestep.    
To evaluate the fusing different modalities in training, 
we consider \textbf{STGCN-3d} and \textbf{STZINB-3d} as our graph neural network baseline and compare two other models: 
(1)~\textbf{STGCN-3d}/\textbf{STZINB-3d} 
are state-of-the-art graph learning frameworks that ingest spatio-temporal urban observations for time series prediction. STZINB particularly focuses on the sparse observations, while we evaluate both models for all five datasets. 
(2)~\textbf{STGCN-3d2d}/\textbf{STZINB-3d2d} employ our homophily-embedding graph structure for informing similarities and localities of neighborhood within cities compared to a standard approach that constructs a graph with spatial adjacency.
(3)~\textbf{STGCN-3d2d1d}/\textbf{STZINB-3d2d1d} adds 1D weather feature that covers temporal information at the city. 
As weather events can potentially influence human behavior~\cite{corcoran2022weather}, we integrate the 1D information as extra dimension for STGCN and attention-input for STZINB as described in~\autoref{sec:stgcn_arch} and ~\autoref{sec:stzinb_arch2}.

In addition, we add historical average (HA) and Random Forest (RF) to the performance comparison serving as the statistic baseline. 
HA is calculated by averaging the number of occurrences  at the same interval (e.g. HA of CDP Ride results of the hourly averages) from the historical observation to predict one time-step ahead. 
RF is trained only on a time series data at each census tract without spatial information, serving as a baseline of machine learning model relying only on time series information.

All trainings of graph neural networks are conducted on a single NVIDIA L40S GPU with 46GB of memory. 
We split datasets into train, test, and validation datasets
comprising 70\%, 20\%, and 10\% of the data.
We use Adam optimizer with learning rate \num{0.001} for STGCN, and for \num{0.0001} for STZINBs 
Following an original study~\cite{yu2017spatio}, we apply a decay rate of 0.7 after every 5 epochs for STGCN.

\begin{table}[tbp!]
  \centering
  \caption{Performance comparison of different models. We use seven metrics to evaluate statistical/ML methods, GCN baselines, and GCNs with multi-modalities on five datasets. The best performance is highlighted in bold.} 
  \begin{tiny}      
   \begin{tabular}{>{\RaggedRight}m{1.6cm} | >{\RaggedRight}m{1.9cm} | *{8}{S[table-format=4.3,table-alignment=right]}} 
    \toprule
    Dataset & Metrics & {HA} & {RF} & {STGCN 3d} & {STGCN 3d2d} & {STGCN 3d2d1d} & {STZINB 3d} & {STZINB 3d2d} & {STZINB 3d2d1d} \\
    \midrule
    \textbf{CDP\_Ride\_1h} & MAE (tract) $\downarrow$ & 7.664 & 3.084 &	3.153 &	3.036 & \textbf{2.996} & {-}&{-} &{-} \\
          & MAE (downtown) & 195.617 &	34.350 & 37.294 & 38.502 & \textbf{35.549} & {-} & {-} & {-} \\
          & MPIW 	$\downarrow$ &  {-} &  {-} &  {-} &  {-} &  {-} &  {-} &  {-} & \\
          & KL{-}Div. $\downarrow$ & 0.0697 &	0.0314& 0.044 &	0.0541 & \textbf{0.0309} &{-}&{-}&{-}\\
          & True{-}zero rate 	$\uparrow$ & \textbf{0.182} & 0.170 & 0.172 & 0.146 & 0.172 &{-}&{-}&{-}\\
          & PICP $\rightarrow 80\%$ & {-}& {-}& {-}& {-}& {-}& {-}& {-}&{-} \\
          & F1 $\uparrow$ & 0.239 & 0.284 & 0.271 & 0.295 & \textbf{0.295} &{-}&{-}&{-}\\
    \midrule
    \textbf{CDP\_Crime\_4h}  & MAE (tract) & 0.216 & 0.228 & 0.213 & 0.211 & 0.211 & 0.162 & 0.161 & \textbf{0.159} \\
          & MAE (downtown) & 0.946 &	0.862 &	0.857 &	0.853 &	\textbf{0.833} &	1.134 &	1.168 &	0.991 \\
          & MPIW & {-}  & {-}	& {-} & {-} & {-} & 0.451 & \textbf{0.446} & 0.450 \\
          & KL{-}Div. & 0.255 & 0.186 & 0.261 & 0.258 & 0.253 & \textbf{0.047} & 0.049 & 0.049 \\
           & PICP & {-} & {-} & {-} & {-} & {-} & 0.954 & \textbf{0.953} & 0.954 \\
          & True{-}zero rate & \textbf{0.884} & 0.878 & 0.883 & 0.883 & 0.883 & 0.852 & 0.853  & 0.853\\
          & F1 & 0.884 & 0.880 & 0.884 & 0.884 & \textbf{0.884} & 0.858 & 0.860 & 0.860 \\
    \midrule
    \textbf{CDP\_Crash\_4h} & MAE (tract)& 0.116 & 0.119 & 0.111 & 0.111 & 0.103 & 0.076 &	0.075 &	\textbf{0.073} \\
          & MAE (downtown) & 0.607 & 0.619 &	0.538 &	0.554 &	\textbf{0.543} & 0.712 &	0.678 &	0.582 \\
          & MPIW &{-} & {-} &	{-} &	{-} &	{-} &	0.161 &	0.165 &	\textbf{0.146}  \\
          & KL{-}Div. & \textbf{0.001} & 0.135 & \textbf{0.001} & \textbf{0.001} & \textbf{0.001} & 0.042 & 0.042 & 0.050 \\
          & PICP & {-} & {-} & {-} & {-} & {-} & 0.959 & 0.960 & \textbf{0.959} \\
          & True{-}zero rate & \textbf{0.940} & 0.938 & \textbf{0.940} & \textbf{0.940} & \textbf{0.940} & 0.928 & 0.928 & 0.930 \\
          & F1 & \textbf{0.940} & 0.939 & \textbf{0.940} & \textbf{0.940} & \textbf{0.940} & 0.929 &	0.930 &	0.932 \\
    \midrule
    \textbf{WPRDC\_Crime\_1D} & MAE (tract) & 0.105 & 0.128 & 0.110 & 0.111 & 0.117 & 0.084 & 0.083 &\textbf{0.082} \\
          & MAE (downtown) & 0.803 & 0.893 &	0.784 &	\textbf{0.779} &	0.785 &	0.886 &	0.882 &	0.863 \\
          & MPIW & {-} &	{-}&	{-}&	{-}&	{-}&	\textbf{0.212} &	0.214 &	0.221  \\
          & KL{-}Div. & 0.062 & 0.048 & 0.098 & 0.071 & 0.062 & \textbf{0.013} & 0.013 &	0.014 \\
          & PICP & {-} & {-} & {-} & {-} & {-} & 0.976 & \textbf{0.976} & 0.977 \\
          & True{-}zero rate & 0.945 & 0.942 & \textbf{0.946} & 0.945 & 0.945 & 0.928 & 0.928 & 0.928 \\
          & F1 & 0.946 & 0.944 & 0.946 & \textbf{0.946} & 0.946 & 0.931 & 0.931 & 0.932 \\
    \midrule
    \textbf{ACODH\_Crime\_6h}& MAE (area) & 0.053 &	0.059 & 0.055 & 0.054 &0.054 & \textbf{0.043} & 0.043 & 0.044 \\
          &MAE (downtown)& 0.786 & 0.881 & 0.781 & 0.777 & 0.772 & 0.759 & \textbf{0.756} & 0.775 \\
          & MPIW & {-} & {-} & {-} & {-} & {-} & \textbf{0.106} & 0.108 &	0.110 \\
          & KL{-}Div. & \textbf{0.0002}	& 0.012 & 0.050 &	0.095 & 0.063 & 0.004 & 0.004 &	0.003 \\
          & PICP	&  {-} &	{-} &	{-} &	{-} &	{-} & \textbf{0.991} & 0.991 & 0.991 \\
          & TZ Ratio & 0.979 & 0.975 & 0.979 & 0.979 & \textbf{0.979} & 0.970 & 0.970 & 0.970 \\
          & F1 & \textbf{0.979} & 0.976 & 0.979 & 0.979 & 0.979 & 0.972 & 0.972 & 0.971 \\
    \bottomrule
  \end{tabular}
  \label{tab:performance}
  \end{tiny}
\end{table}
\subsection{Model Comparison \& Ablation Study}\label{sec:model_comp}

\autoref{tab:performance} compares results of eight performance metrics on five datasets.
The table also serves as the ablation study of how adding additional modality helps model learning. 
The bold entry indicates the best performance for each metrics calculated on each observation dataset. 
Note that STZINBs do not converge during the training, reaming blank in the table.  

Four of five case studies show that multi-modality input outperform the graph neural networks with 3D only input.
MAE in CDP Ride reduces by 4.9\% in all tracts and 4.7\% in downtown.
Results of CDP Crime shows that STZINB-3d2d1d outperforms other models in MAE over all tracts for taking into consideration the sparsity of data. 
But, in downtown where crimes occurs more frequent, STGCN-3d2d1d performs better.
We observe the similar trend in CDP Crash and WPRDC Crime, where the zero-inflated model with additional modalities results in the best performance in MAE over the entire city, while STGCN may show competitive performance in the prediction task in the downtown.
We also observe that the STZINB model consistently exhibits improved performance with the addition of both 2D homophily information and the combination of 2D homophily and 1D trend information

In contrast, the homophily-embedding graph structure may not always result in the best performance. 
MAE over the entire Oakland from ACODH-crime dataset shows that conventional STZINB-3d shows the best performance and adding modalities do not significantly reduce the error rate. 
This limited impact of spatial information on crime prediction in Oakland may be due to the concentrated nature of crime patterns, with only a few locations experiencing high crime rates. 
Additionally, due to the Oakland's mild climate, the weather-related features may not significantly contribute to improving model performance.
Overall, adding 2D and 1D modalities via our heterogeneous data framework allows models to capture urban information, improving the prediction capabilities.

\begin{figure}[h]
    \centering
    \begin{minipage}{.49\columnwidth}
        \subfloat[Entire tracts\label{fig:chicago_rideshare_time}]{\includegraphics[clip, width=1\columnwidth]{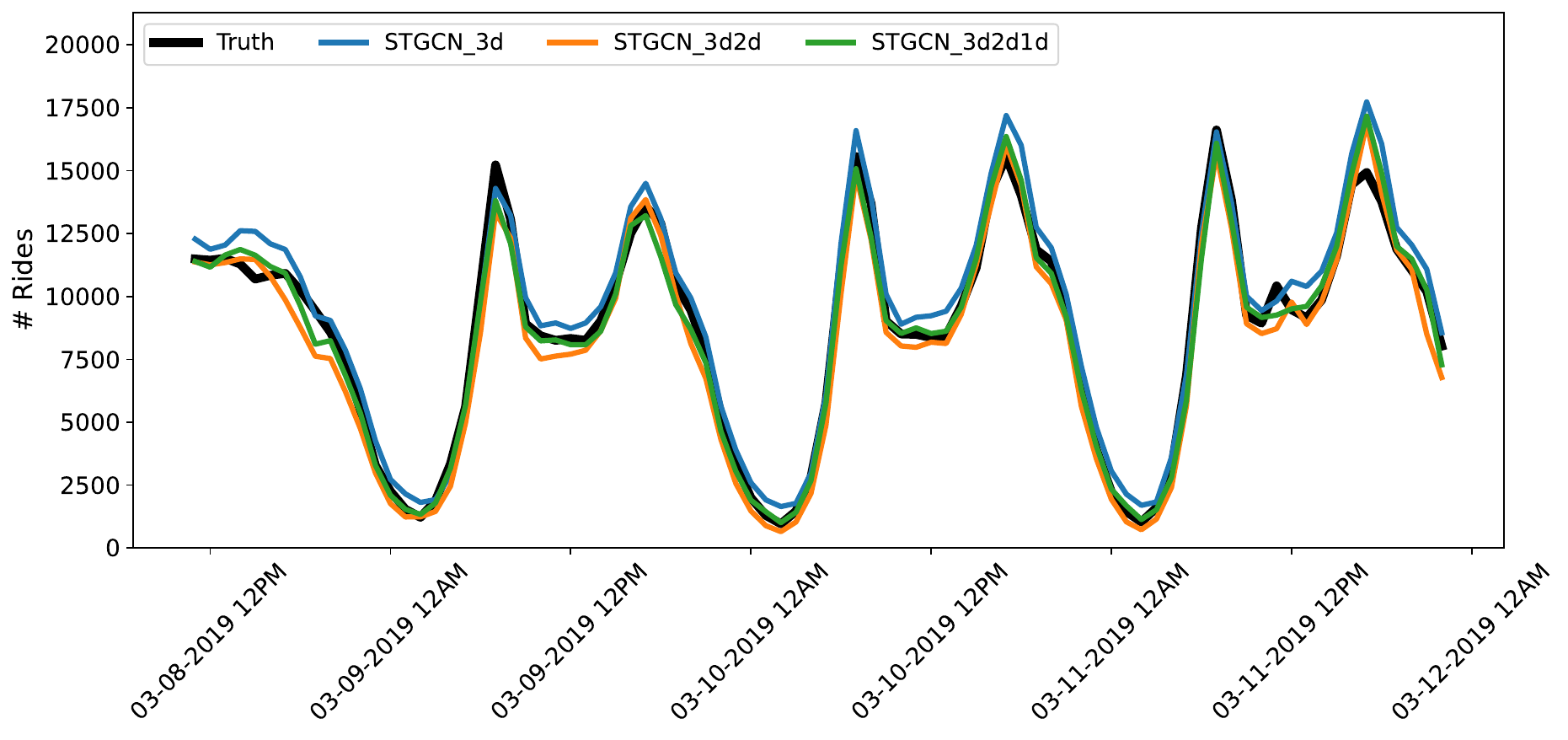}}
    \end{minipage}
    \begin{minipage}{.49\columnwidth}
        \subfloat[Highest demand tract\label{fig:chicago_rideshare_time_downtown}]{\includegraphics[clip, width=1\columnwidth]{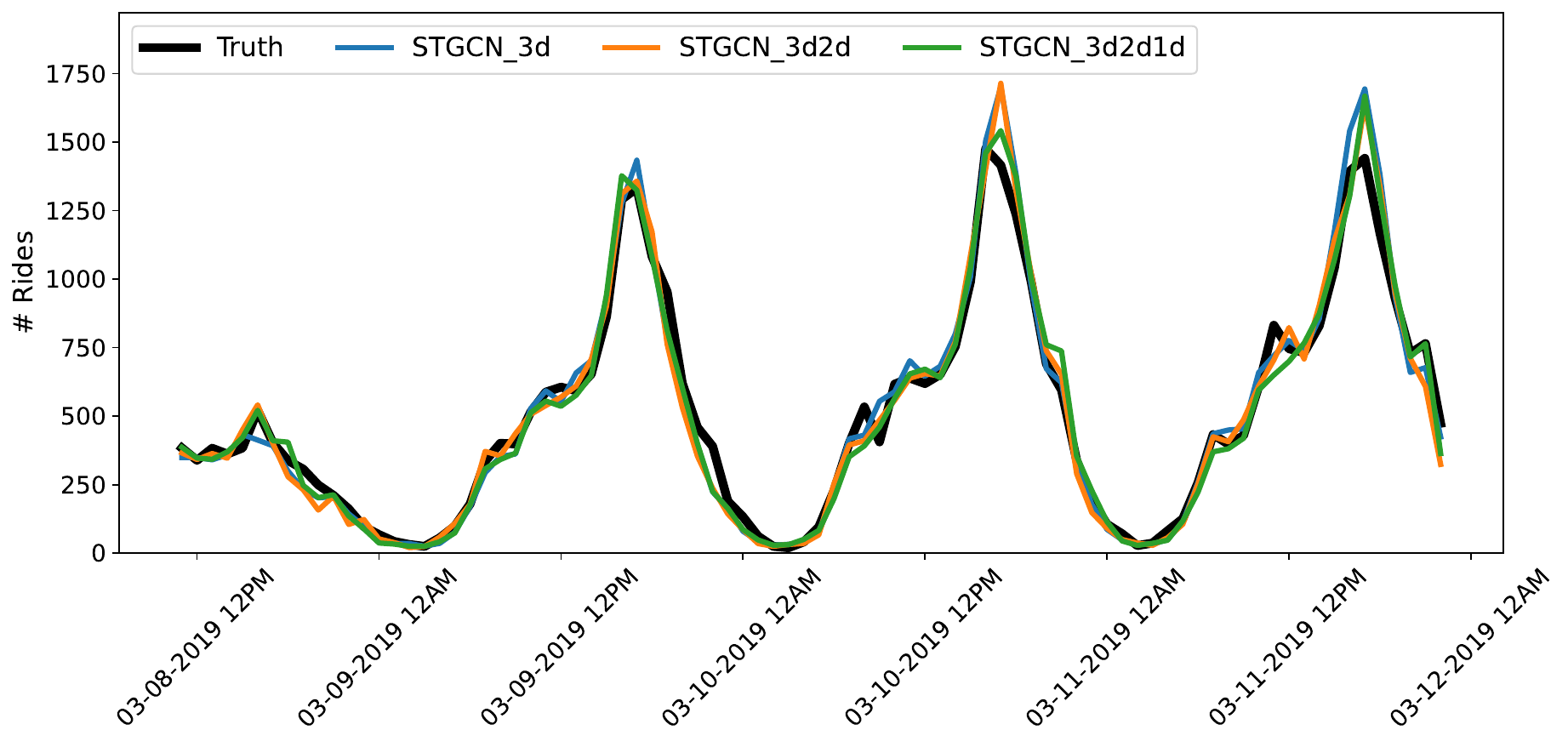}}
    \end{minipage}
    \caption{Results of ride-sourcing demand predictions from STGCNs (a) over the entire Chicago and (b) at the highest demand area, between 11AM March 8th, 2019 to 10PM March 12th, 2019.}
    \label{fig:chi_ride_timeseries}
\end{figure}
\begin{figure}[h]
    \centering
    \begin{minipage}{.49\columnwidth}
        \subfloat[CDP Crime\label{fig:time_chicago_crime}]{\includegraphics[clip, width=1\columnwidth]{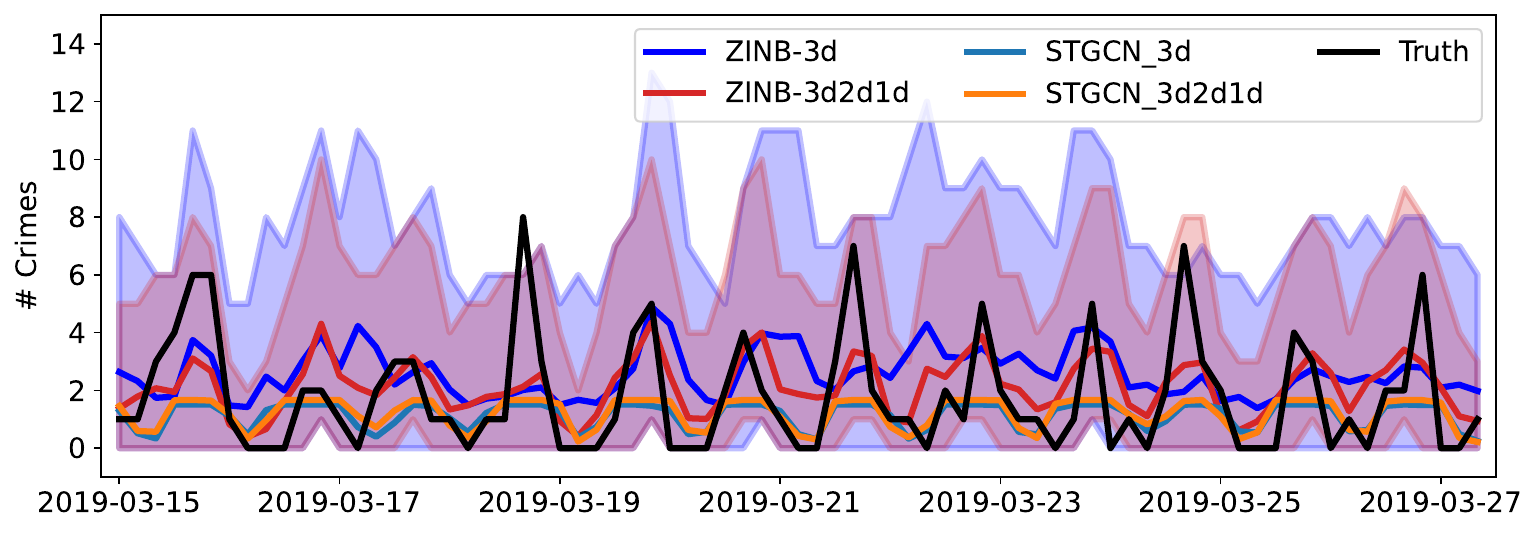}}
    \end{minipage}
    \begin{minipage}{.49\columnwidth}
        \subfloat[CDP Crash\label{fig:time_chicago_crash}]{\includegraphics[clip, width=1\columnwidth]{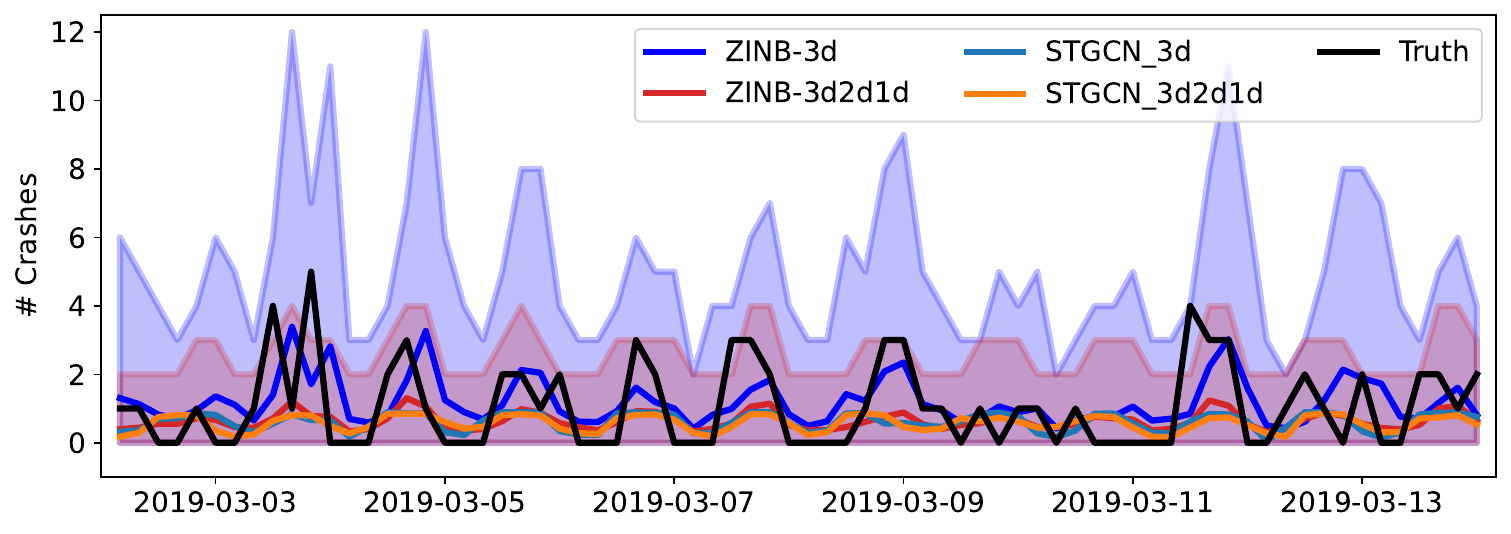}}
    \end{minipage}
    \begin{minipage}{.49\columnwidth}
        \subfloat[WPRDC Crime\label{fig:time_pts_crime}]{\includegraphics[clip, width=1\columnwidth]{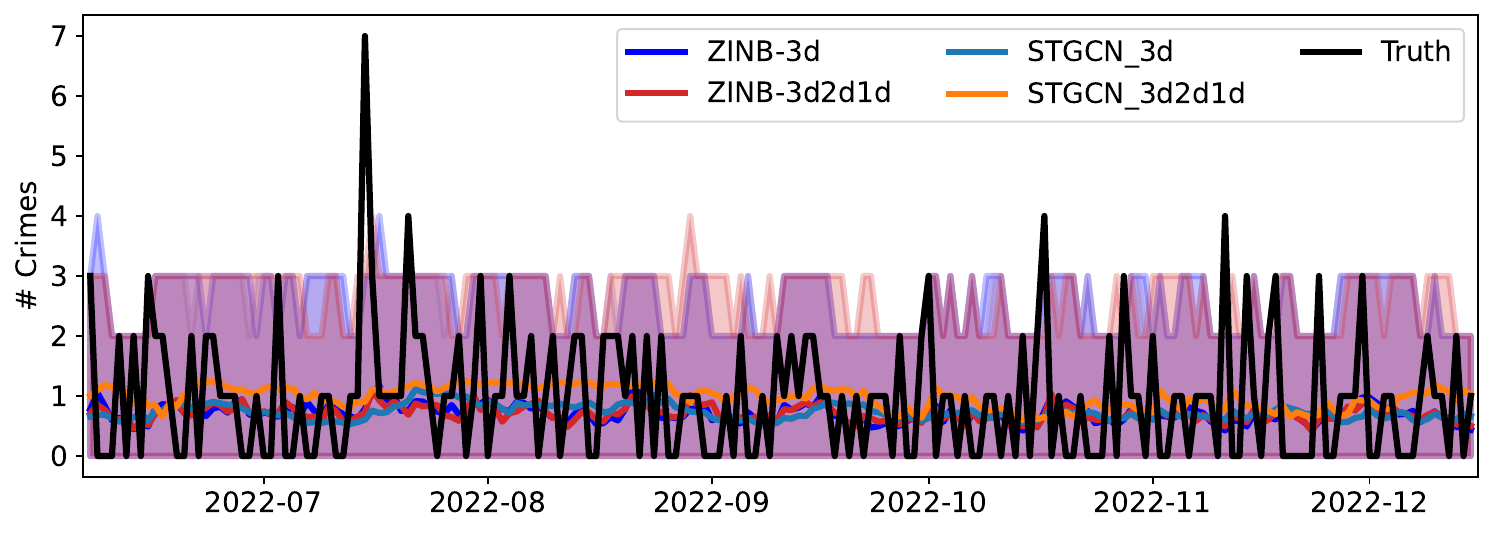}}
    \end{minipage}
    \begin{minipage}{.49\columnwidth}
        \subfloat[ACODH Crime\label{fig:time_oak_crime}]{\includegraphics[clip, width=1\columnwidth]{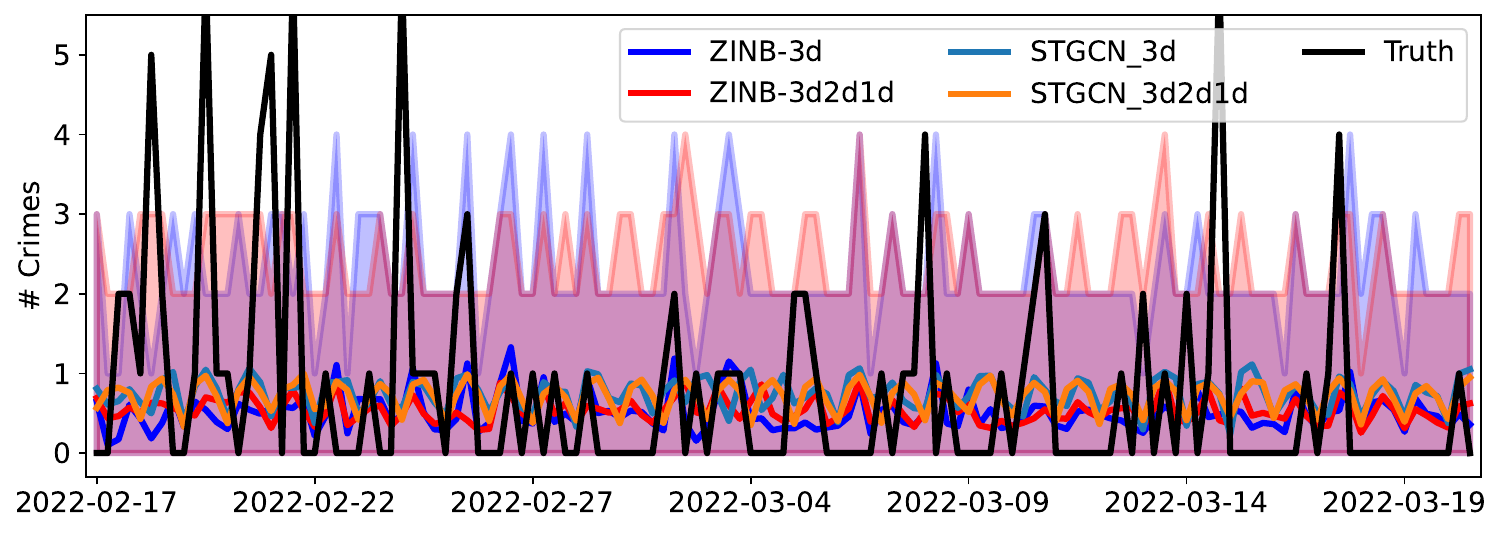}}
    \end{minipage}
    \caption{Plots for prediction results from GCNs over selected time series. Similar to~\autoref{fig:chicago_rideshare_time_downtown}, solid lines represent the occurrences at a tract with the highest observation. (adding shade and solid line exp.)}
    \label{fig:timeseries_downtown2}
\end{figure}

\subsection{Case Study: Evaluating Temporal Performance}\label{sec:time}

To get insight of how model prediction and uncertainty ranges evolve over time whether adding modalities, we visualize the time series results of five datasets between the 3D baseline and the other multi-modal models.

\autoref{fig:chi_ride_timeseries} plots the number of ride demands and the predictions from March 8th, 2019 11PM to March 12th, 2019 11PM in the entire Chicago~\autoref{fig:chicago_rideshare_time} and a tract with the highest ride-sourcing demands~\autoref{fig:chicago_rideshare_time_downtown}. 
The time series plots suggest that incorporating 2D homophily information may also enable the model to better capture temporal dynamics, particularly during periods of rapid demand fluctuation. 
In ride-sourcing case, 1D weather information adjust the model prediction much closer to ground-truth but when demands changes rapidly, it may smooth the predictions.

\autoref{fig:timeseries_downtown2} visualizes the time series of the number of CDP crash and crimes in three cities with the highest records across three different cities at randomly selected time periods.
Solid lines represent the deterministic from STGCNs and mean predictions from STZINBs. 
The shaded area quantifies the uncertainty range (10\% to 90\% interval) from STZINBs.
The STZINB models demonstrate its effectiveness by generating uncertainty ranges that closely follow the ground truth for both crime and traffic crashes in Chicago by additional modalities, and by accurately replicating observed spikes in the data. 
The STGCN model's predictions fall short to replicating the observed patterns, particularly in its inability to accurately predict values exceeding a certain value (i.e., exhibiting a performance plateau).
This highlights the limitations of STGCN for sparse observations in contrast to the ride-sourcing dataset. 

However, we also observe some inherent limitations in the exact model predictions as shown in~\autoref{fig:time_pts_crime} and~\autoref{fig:time_oak_crime} when relying solely on mean prediction values derived from both GCNs in Oakland and Pittsburgh. 
The analysis shows that while the uncertainty ranges for ACODH crime data in Oakland often capture high crime spikes, they generally overestimate the true values, even when crime occurrences are zero.
These results suggests that further development is required for the improvement at our future work.

\begin{figure}[htbp!]
    \centering
    \begin{minipage}{.48\columnwidth}
        \subfloat[CDP Ride\label{fig:chicago_rideshare_diff}]{\includegraphics[clip, width=1\columnwidth]{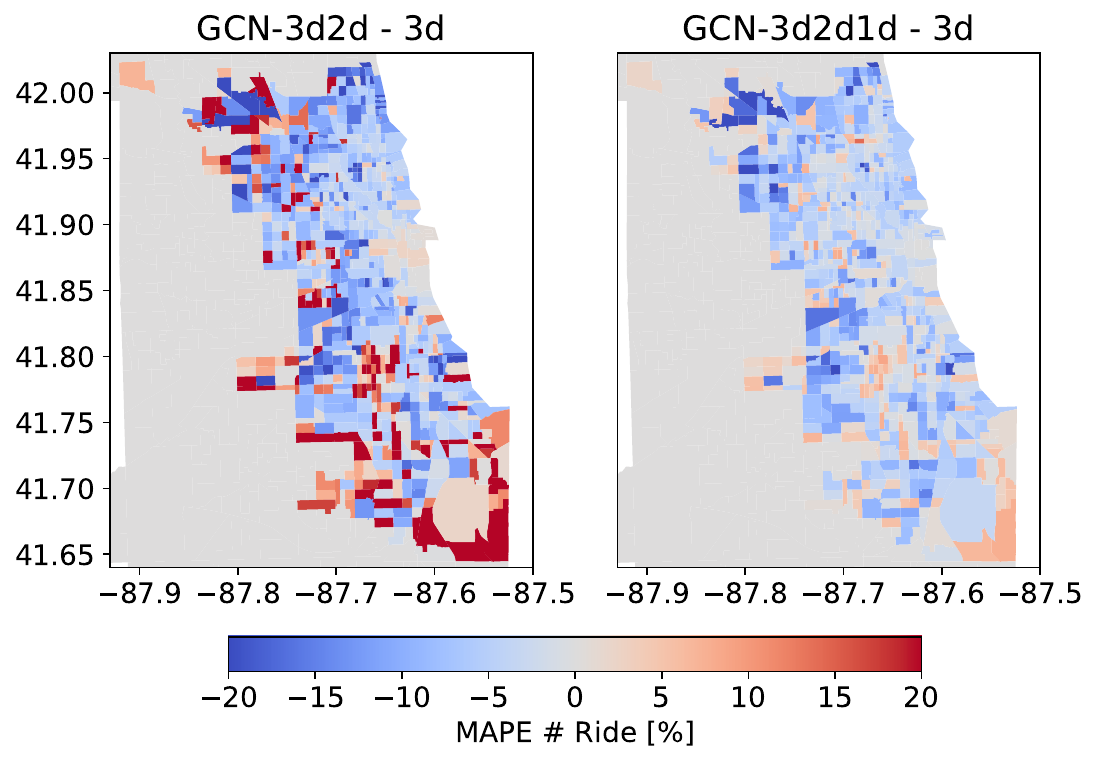}}
    \end{minipage}
    \begin{minipage}{.48\columnwidth}
        \subfloat[CDP Crime\label{fig:chicago_crime_diff}]{\includegraphics[clip, width=1\columnwidth]{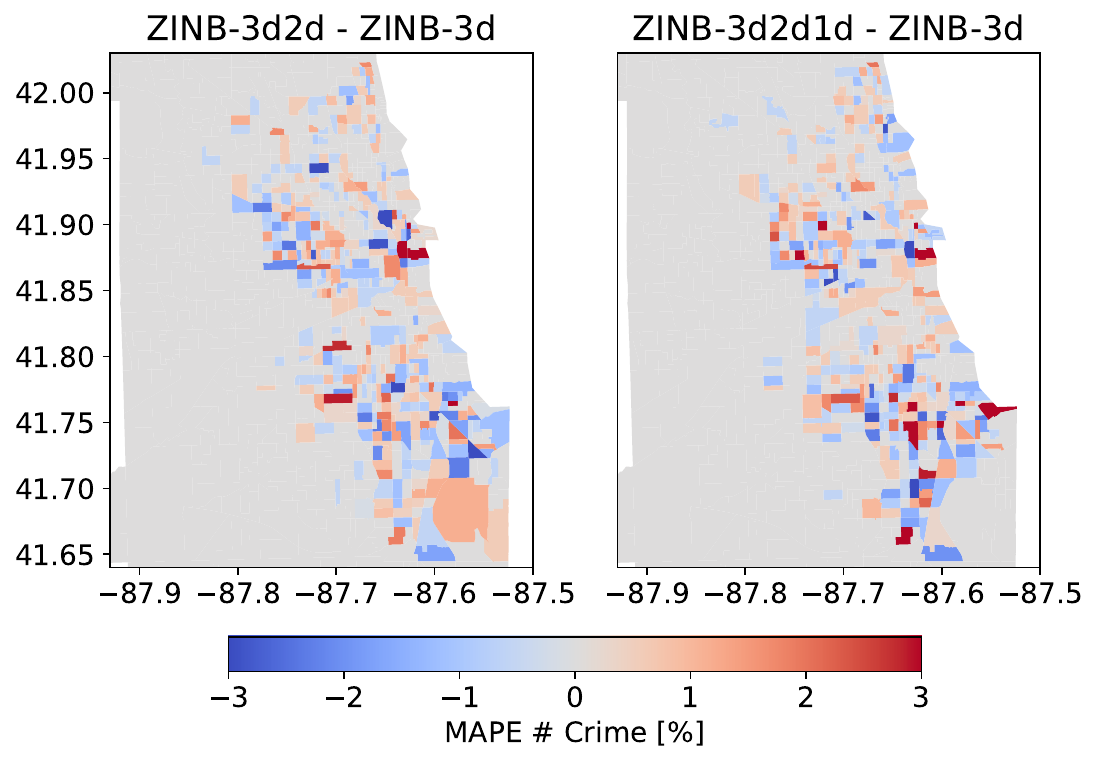}}
    \end{minipage}
    \begin{minipage}{.48\columnwidth}
        \subfloat[CDP Crash\label{fig:chicago_crash_diff}]{\includegraphics[clip, width=1\columnwidth]{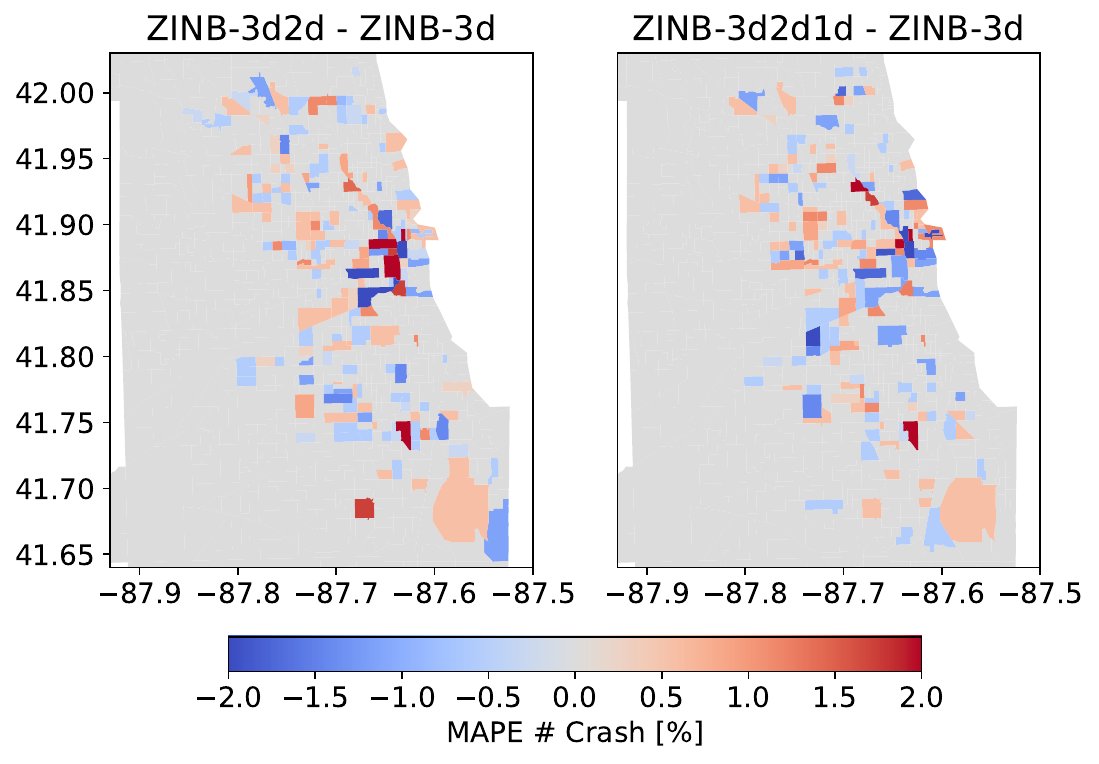}}
    \end{minipage}
    \caption{
    Spatial distribution of differences in mean absolute percentage error (MAPE) from multi-modal data learning for Ride-sourcing (top left), Crime (top right), and Crash (bottom) observations in Chicago. Blue regions represents census tracts where the inclusion of 2D or combined 2D and 1D information reduces MAPE compared to time-series-only input. Red regions highlight tracts where multi-modal integration do not help improving predictions.
    }
    \label{fig:chi-mape-diff}
\end{figure}

\begin{figure}[htbp!]
    \centering
    \begin{minipage}{.49\columnwidth}
        \subfloat[WPRDC Crime\label{fig:pts_crime_diff}]{\includegraphics[clip, width=1\columnwidth]{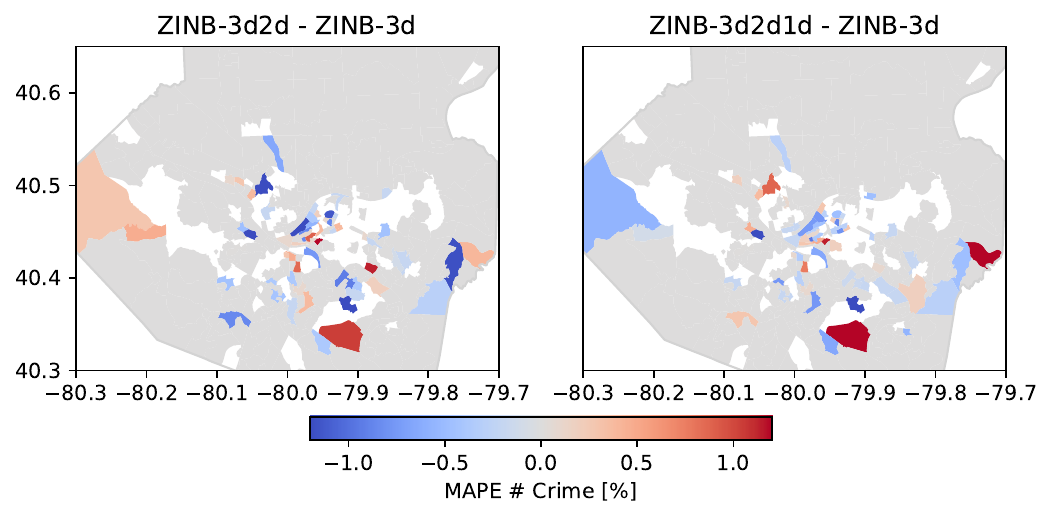}}
    \end{minipage}
    \begin{minipage}{.49\columnwidth}
        \subfloat[ACODH Crime\label{fig:oak_crime_diff}]{\includegraphics[clip, width=1\columnwidth]{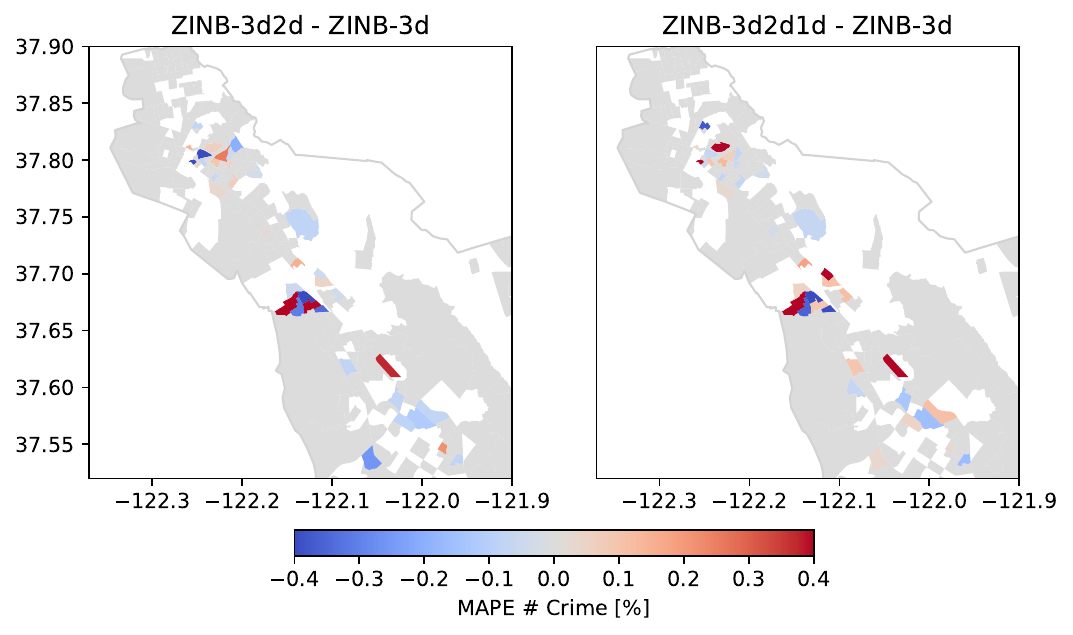}}
    \end{minipage}
    
    \caption{In the same way to~\autoref{fig:chi-mape-diff}, visualizing spatial distribution of differences in MAPE from multi-modal data learning for crime observation in (a) Pittsburgh and (b) Oakland. Tracts colored white represent areas with consistently infrequent crime occurrences ($\leq$ 1 throughout the test period). }
    \label{fig:crime-mape-diff}
\end{figure}

\subsection{Spatial Impact on Multi-Modality Data-Fusion}
In this section, we conduct a spatial analysis of model performance to evaluate the impact of multi-modality fusion on the percentage error at the census tract level. 
Specifically, we examine the spatial distribution of error differences observed with and without the incorporation of those additional modalities.
We visualize the difference in the mean absolute percentage error (MAPE)~\autoref{eq:mape} between the 3D+2D+1D and 3D models as well as between 3D+2D and 3D models in Chicago~(\autoref{fig:chi-mape-diff}) and the other two cities~(\autoref{fig:crime-mape-diff}). 
Blue regions indicate census tracts where the inclusion of additional modalities leads to a reduction in prediction errors, while red ones represent opposite impact. 
We present results from STGCNs for ride-sourcing case in Chicago, while we evaluate results from STZINBs for other cases. 

\autoref{fig:chicago_rideshare_diff} shows that adding 2D homophily-embedding graph reduces errors in 38.4\% of tracts and adding 1D weather to the homophily-embedding graph learning helps 42.8\% of tracts. 
Particularly, tracts surrounding the central Chicago area show more significant error reductions by 10\% up to 20\%. 
Adding weather feature alleviates the error increment south side and northwest part of Chicago.  
\autoref{fig:chicago_crime_diff} reveals that approximately 13\% of census tracts are benefit to have multi-modality in graph learning for CDP Crime dataset.
It is noteworthy that adding weather feature significantly reduce crime prediction area by 29.2\% at the neighborhood so called `West Loop', one of the newly developed areas in downtown Chicago.  

Results from the CPD Crash case study in ~\autoref{fig:chicago_crash_diff} show that
approximately 7.1\% of census tracts demonstrate a reduction in prediction error when using the 3D+2D model, while 6\% show improvement with the 3D+2D+1D model.
We observe significant associations between the inclusion of weather information and error reduction in census tracts located along the Interstate 55 (I-55) highway, as well as on the west side of downtown Chicago, where Interstates 90 and 290 create a high-traffic area.
The pattern is not shown in solely adding build-in environments, indicating that the temporal features in a city exhibit a strong localized effect on trend variations within smaller geographic areas. 

Compared to Chicago, Pittsburgh and Oakland are formed with smaller and more concentrated populated structures, reflecting that limited number of areas exhibits the spatial error structures in~\autoref{fig:crime-mape-diff}.
Note that due to data sparsity, there are limitations in performance improvements that reflect the predominantly gray areas observed in~\autoref{fig:crime-mape-diff}.
We observe that adding 2D modality to graph creation reduces errors at 11.4\% of tracts and combination of 2D and 1D contribute by 13.0\% of tracts shown in~\autoref{fig:pts_crime_diff}.
The results indicate that the southern downtown Pittsburgh along the Allegheny River and the eastern end of Allegheny County have a notable improvement in performance due to the inclusion of additional modalities in the graph learning framework. 

%
In the Oakland crime prediction task, 
the incorporation of 2D homophily information and the combination of 2D homophily and 1D weather trend information resulted in a reduction of prediction error in 9.6\% of census tracts and 7.6\% of census tracts, respectively. 
Despite the relatively small percentage differences, it is noteworthy that the incorporation of an extensive set of heterogeneous built environment features consistently reduces crime error rates in Fremont and Union City, located in the southern part of Alameda County (as shown in \autoref{fig:oak_crime_diff}). 
Furthermore, we observe a similar pattern of error reduction in the eastern and western neighborhoods of Hayward, located in the center of the map (see~\autoref{fig:oak_crime_diff}).
However, adding the weather information results in increasing errors at 9.2\% of census tracts, suggesting that temporal weather information may not help the error reduction in California, which has mild weather annually.  

\label{sesc:experiment}

\section{CONCLUSION}
We propose a heterogeneous multi-modal spatio-temporal graph learning framework that is easily applicable to support various prediction tasks with various modalities.
This framework presents a novel and efficient graph learning approach that addresses the need for manual data curation and iterative fine-tuning typically required for each locale and problem by conventional methods.
We demonstrate that our framework is minimum to adjust but capable of incorporating temporal information over the entire city as well as complex urban local characteristics to graph learning. 
Experiments exhibit a consistent improvement in performance across four of the five prediction tasks.
While 3D spatio-temporal data can provide reasonable prediction accuracy, the integration of 2D (spatial) and 1D (temporal) data sources enhances the predictive performance, providing deeper insights to interactions of urban environment. 

\label{sec:discussion}

\appendix
\section{Appendix}
\subsection{Homophily-Embedded Adjacency Matrix}\label{append:homo}

\begin{figure}[h]
    \centering
    \begin{minipage}{.49\columnwidth}
    \subfloat[Pittsburgh\label{fig:adj_weight_pts}]{\includegraphics[clip, width=1\columnwidth]{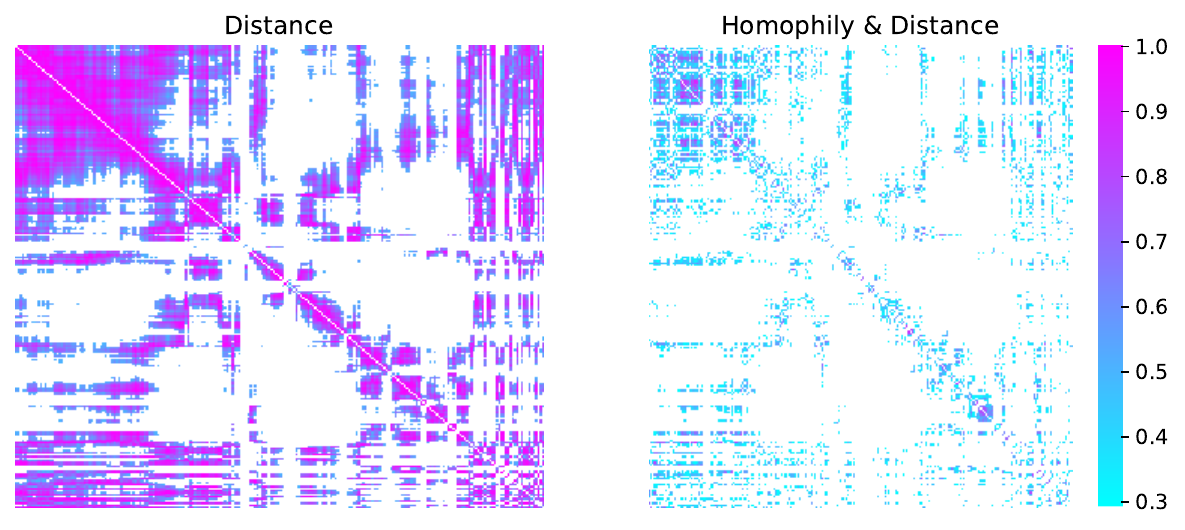}}
    \end{minipage}
    \begin{minipage}{.49\columnwidth}
        \subfloat[Oakland\label{fig:adj_weight_oak}]{\includegraphics[clip, width=1\columnwidth]{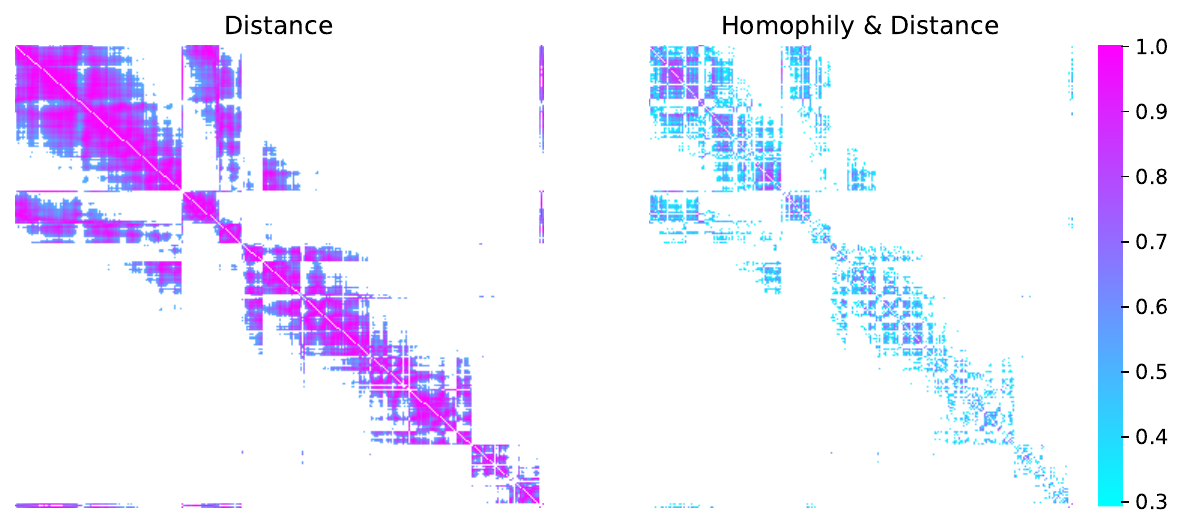}}
    \end{minipage}
    \caption{Similar to~\autoref{fig:adj_weight}, we visualize heatmaps of the weights of adjacency matrix from distance-based and homophily-embedding approaches in (a)~Pittsburgh and (b)~Oakland.}
    \label{fig:adj_weight_append}
\end{figure}

We display how our homophily-embedding approach modifies the weights of the graph during graph learning for the Pittsburgh~(\autoref{fig:adj_weight_pts}) and Oakland~(\autoref{fig:adj_weight_oak}) cases.

\subsection{Details of Feature Dataset}\label{append:features}
The comprehensive 48 features acquired from five source of publicly available dataset and the descriptions are listed in~\autoref{tab:full_variable_names}.

\begin{table}[b]
\centering
\scriptsize
\caption{Comprehensive list of all features calculating correlations among census tracts.} 
\begin{tabular}{p{.75cm} | p{1.6cm}p{6cm}p{1.2cm}}
\toprule
Category & Variable  &  Description & Data source  \\
\midrule
Demo. & totpop & total population & USCB\\
  & popden & population density per square mile & "  \\
  & pctmale & percentage of male population & "  \\
  & hhsize & average household size & "  \\
  & pcthighschool & percentage of high school diploma& "  \\
  & pctsomecollege & percentage of associate degree & "  \\
  & pctbachelor & percentage of bachelor or higher degree& "  \\
  & medage & median of age & "  \\
  & pctyoung & percentage of population between 18-34 & "  \\
  & pctmiddleyoung & percentage of population between 18-44 & "  \\
  & pctasian & percentage of asian population & "  \\
  & pctwhite & percentage of white population & "  \\
  & pctblack & percentage of black population & "  \\
  & pcthisp & percentage of hispanic population & "  \\
  & carown & percentage of households with at least one car & "  \\
  & pct2car & percentage of households with 2 or more cars& "  \\
  & timetowork & mean commute time to workplace & "  \\
  & pcttransit & percentage of workers taking transit to work & "  \\
  & pctdrialone & percentage of workers driving alone to work & "  \\
  & numworker & number of workers & "  \\
  & unemploy & percentage of unemployment & "  \\
  & medhhinc & median household income & "  \\
  & incpercap & income per capita & "  \\
  & pctpoverty & percentage of individuals below poverty & "  \\
  & pctlowinc & percentage of low-income population (\$25k less) & "  \\
  & pctmodinc & percentage of moderate-income population (\$25k-\$50k) & "  \\
  & pctlowmidinc & percentage of lower middle-income population (\$50-\$75k less) & "  \\
  & pcthighmidinc & percentage of higer middle-income population (\$75-\$100k) & "  \\
  & pctmidinc & percentage of middle-income population (\$50k-\$100k) & "  \\
  & pcthighinc & percentage of high-income population (\$100k or more) & "  \\
  & giniindex & gini index & "  \\
  & pctrentocc & percentage of renter-occupied housing units& "  \\
  & pctdesinfam & percentage of detached single-family & "  \\
  & pctsinfam & percentage of single-family & "  \\
  & medvalue & median house value& "  \\
  & medrent & median rent value & "  \\ \midrule
 Econ. & Retail & percentage of employments in retail trade& LEHD/USCB  \\
  & Office & percentage of employments in office work & "  \\
  & Service & percentage of employments in service industry & "  \\
  & Entertain & percentage of employments in entertainment industry & "  \\
  & Indus & percentage of employments in other industries& "  \\ \midrule
 Road & RdNetwkDen & density of roads per square mile & OpenStreetMap  \\
  & InterstDen & density of intersections per square mile & "  \\
  & Walkscore & walk scire index & Walkscore.com  \\ \midrule
 Land & usgs\_water & open water area& USGS \\
  & usgs\_developed & developed area & " \\
  & usgs\_cultivated & cultivated crops area & " \\
  & usgs\_vegetation & vegetation area (e.g., forest, grass land)  & " \\  
\bottomrule
\end{tabular}
\parbox[t]{0.98\columnwidth}
{\vskip3pt{\footnotesize $^*$Note: LEHD: Longitudinal Employer-Household Dynamics; USGS: United States Geological Survey. USCB: United States Census Bureau}}
\label{tab:full_variable_names}
\end{table}\label{sec:appendix}
\clearpage
\bibliographystyle{unsrtnat}
\bibliography{references}

@article{corcoran2022weather,
  title={Weather and crime: a systematic review of the empirical literature},
  author={Corcoran, Jonathan and Zahnow, Renee},
  journal={Crime Science},
  volume={11},
  number={1},
  pages={16},
  year={2022},
  publisher={Springer},
  doi={https://doi.org/10.1186/s40163-022-00179-8}
}

@article{stubbs2018searching,
  title={Searching for safety: crime prevention in the era of Google},
  author={Stubbs-Richardson, Megan S and Cosby, Austin K and Bergene, Karissa D and Cosby, Arthur G},
  journal={Crime Science},
  volume={7},
  pages={1--13},
  year={2018},
  publisher={Springer},
  doi={https://doi.org/10.1186/s40163-018-0095-3}
}

@article{rayhan2023aist,
  title={Aist: An interpretable attention-based deep learning model for crime prediction},
  author={Rayhan, Yeasir and Hashem, Tanzima},
  journal={ACM Transactions on Spatial Algorithms and Systems},
  volume={9},
  number={2},
  pages={1--31},
  year={2023},
  publisher={ACM New York, NY},
  doi={https://dl.acm.org/doi/10.1145/3582274}
}

@inproceedings{wang2022hagen,
  title={Hagen: Homophily-aware graph convolutional recurrent network for crime forecasting},
  author={Wang, Chenyu and Lin, Zongyu and Yang, Xiaochen and Sun, Jiao and Yue, Mingxuan and Shahabi, Cyrus},
  booktitle={Proceedings of the AAAI Conference on Artificial Intelligence},
  volume={36},
  number={4},
  pages={4193--4200},
  year={2022},
  doi={https://doi.org/10.1609/aaai.v36i4.20338}
}

@inproceedings{sun2021crimeforecaster,
  title={Crimeforecaster: Crime prediction by exploiting the geographical neighborhoods’ spatiotemporal dependencies},
  author={Sun, Jiao and Yue, Mingxuan and Lin, Zongyu and Yang, Xiaochen and Nocera, Luciano and Kahn, Gabriel and Shahabi, Cyrus},
  booktitle={Machine Learning and Knowledge Discovery in Databases. Applied Data Science and Demo Track: European Conference, ECML PKDD 2020, Ghent, Belgium, September 14--18, 2020, Proceedings, Part V},
  pages={52--67},
  year={2021},
  organization={Springer}, 
  doi={https://doi.org/10.1007/978-3-030-67670-4_4}
}

@inproceedings{wang2025uncertainty,
  title={Uncertainty-aware crime prediction with spatial temporal multivariate graph neural networks},
  author={Wang, Zepu and Ma, Xiaobo and Yang, Huajie and Lyu, Weimin and Liu, Yang and Sun, Peng and Guntuku, Sharath Chandra},
  booktitle={ICASSP 2025-2025 IEEE International Conference on Acoustics, Speech and Signal Processing (ICASSP)},
  pages={1--5},
  year={2025},
  organization={IEEE},
  doi={10.1109/ICASSP49660.2025.10889685}
}

@article{gao2024uncertainty,
  title={Uncertainty-aware probabilistic graph neural networks for road-level traffic crash prediction},
  author={Gao, Xiaowei and Jiang, Xinke and Haworth, James and Zhuang, Dingyi and Wang, Shenhao and Chen, Huanfa and Law, Stephen},
  journal={Accident Analysis \& Prevention},
  volume={208},
  pages={107801},
  year={2024},
  publisher={Elsevier},
  doi={https://doi.org/10.1016/j.aap.2024.107801}
}

@inproceedings{li2022spatial,
  title={Spatial-temporal hypergraph self-supervised learning for crime prediction},
  author={Li, Zhonghang and Huang, Chao and Xia, Lianghao and Xu, Yong and Pei, Jian},
  booktitle={2022 IEEE 38th international conference on data engineering (ICDE)},
  pages={2984--2996},
  year={2022},
  organization={IEEE},
  doi={10.1109/ICDE53745.2022.00269},
}

@inproceedings{song2020spatial,
  title={Spatial-temporal synchronous graph convolutional networks: A new framework for spatial-temporal network data forecasting},
  author={Song, Chao and Lin, Youfang and Guo, Shengnan and Wan, Huaiyu},
  booktitle={Proceedings of the AAAI conference on artificial intelligence},
  volume={34},
  number={01},
  pages={914--921},
  year={2020},
  doi={https://doi.org/10.1609/aaai.v34i01.5438}
}

@inproceedings{yan2018spatial,
  title={Spatial temporal graph convolutional networks for skeleton-based action recognition},
  author={Yan, Sijie and Xiong, Yuanjun and Lin, Dahua},
  booktitle={Proceedings of the AAAI conference on artificial intelligence},
  volume={32},
  number={1},
  year={2018},
  doi={ https://doi.org/10.1609/aaai.v32i1.12328}
}

@article{zhang2020graph,
  title={Graph deep learning model for network-based predictive hotspot mapping of sparse spatio-temporal events},
  author={Zhang, Yang and Cheng, Tao},
  journal={Computers, Environment and Urban Systems},
  volume={79},
  pages={101403},
  year={2020},
  publisher={Elsevier},
  doi={https://doi.org/10.1016/j.compenvurbsys.2019.101403}
}

@article{yang2021tctn,
  title={TCTN: A 3D-temporal convolutional transformer network for spatiotemporal predictive learning},
  author={Yang, Ziao and Yang, Xiangrui and Lin, Qifeng},
  journal={arXiv preprint arXiv:2112.01085},
  year={2021}
}

@article{tang2022short,
  title={Short-term load forecasting using channel and temporal attention based temporal convolutional network},
  author={Tang, Xianlun and Chen, Hongxu and Xiang, Wenhao and Yang, Jingming and Zou, Mi},
  journal={Electric Power Systems Research},
  volume={205},
  pages={107761},
  year={2022},
  publisher={Elsevier},
  doi={https://doi.org/10.1016/j.epsr.2021.107761}
}

@article{lin2024channel,
  title={Channel attention \& temporal attention based temporal convolutional network: A dual attention framework for remaining useful life prediction of the aircraft engines},
  author={Lin, Lin and Wu, Jinlei and Fu, Song and Zhang, Sihao and Tong, Changsheng and Zu, Lizheng},
  journal={Advanced Engineering Informatics},
  volume={60},
  pages={102372},
  year={2024},
  publisher={Elsevier},
  doi={https://doi.org/10.1016/j.aei.2024.102372}
}

@article{li2017diffusion,
  title={Diffusion convolutional recurrent neural network: Data-driven traffic forecasting},
  author={Li, Yaguang and Yu, Rose and Shahabi, Cyrus and Liu, Yan},
  journal={arXiv preprint arXiv:1707.01926},
  year={2017}
}

@article{yu2017spatio,
  title={Spatio-temporal graph convolutional networks: A deep learning framework for traffic forecasting},
  author={Yu, Bing and Yin, Haoteng and Zhu, Zhanxing},
  journal={arXiv preprint arXiv:1709.04875},
  year={2017}
}

@article{liu2020urban,
  title={Urban big data fusion based on deep learning: An overview},
  author={Liu, Jia and Li, Tianrui and Xie, Peng and Du, Shengdong and Teng, Fei and Yang, Xin},
  journal={Information Fusion},
  volume={53},
  pages={123--133},
  year={2020},
  publisher={Elsevier},
  doi={https://doi.org/10.1016/j.inffus.2019.06.016}
}

@article{zheng2015methodologies,
  title={Methodologies for cross-domain data fusion: An overview},
  author={Zheng, Yu},
  journal={IEEE transactions on big data},
  volume={1},
  number={1},
  pages={16--34},
  year={2015},
  publisher={IEEE},
  doi={10.1109/TBDATA.2015.2465959}
}

@article{zou2025deep,
  title={Deep learning for cross-domain data fusion in urban computing: Taxonomy, advances, and outlook},
  author={Zou, Xingchen and Yan, Yibo and Hao, Xixuan and Hu, Yuehong and Wen, Haomin and Liu, Erdong and Zhang, Junbo and Li, Yong and Li, Tianrui and Zheng, Yu and others},
  journal={Information Fusion},
  volume={113},
  pages={102606},
  year={2025},
  publisher={Elsevier},
  doi={https://doi.org/10.1016/j.inffus.2024.102606}
}

@article{fadhel2024comprehensive,
  title={Comprehensive systematic review of information fusion methods in smart cities and urban environments},
  author={Fadhel, Mohammed A and Duhaim, Ali M and Saihood, Ahmed and Sewify, Ahmed and Al-Hamadani, Mokhaled NA and Albahri, AS and Alzubaidi, Laith and Gupta, Ashish and Mirjalili, Sayedali and Gu, Yuantong},
  journal={Information Fusion},
  pages={102317},
  year={2024},
  publisher={Elsevier},
  doi={https://doi.org/10.1016/j.inffus.2024.102317}
}

@article{zhang2022multi,
  title={Multi-modal graph interaction for multi-graph convolution network in urban spatiotemporal forecasting},
  author={Zhang, Lingyu and Geng, Xu and Qin, Zhiwei and Wang, Hongjun and Wang, Xiao and Zhang, Ying and Liang, Jian and Wu, Guobin and Song, Xuan and Wang, Yunhai},
  journal={Sustainability},
  volume={14},
  number={19},
  pages={12397},
  year={2022},
  publisher={MDPI},
  doi={ https://doi.org/10.3390/su141912397}
}

@article{zhang2022machine,
  title={Machine learning approach for spatial modeling of ridesourcing demand},
  author={Zhang, Xiaojian and Zhao, Xilei},
  journal={Journal of Transport Geography},
  volume={100},
  number={C},
  year={2022},
  publisher={Elsevier},
  doi={https://doi.org/10.1016/j.jtrangeo.2022.103310}
}

@article{zhang2024travel,
  title={Travel demand forecasting: A fair ai approach},
  author={Zhang, Xiaojian and Ke, Qian and Zhao, Xilei},
  journal={IEEE Transactions on Intelligent Transportation Systems},
  year={2024},
  publisher={IEEE},
  doi={10.1109/TITS.2024.3395061}
}

@article{ben1996travel,
  title={Travel demand model system for the information era},
  author={Ben-Akivai, Moshe and Bowman, John L and Gopinath, Dinesh},
  journal={Transportation},
  volume={23},
  pages={241--266},
  year={1996},
  publisher={Springer}
}

@inproceedings{zhuang2022uncertainty,
  title={Uncertainty quantification of sparse travel demand prediction with spatial-temporal graph neural networks},
  author={Zhuang, Dingyi and Wang, Shenhao and Koutsopoulos, Haris and Zhao, Jinhua},
  booktitle={Proceedings of the 28th ACM SIGKDD Conference on Knowledge Discovery and Data Mining},
  pages={4639--4647},
  year={2022},
  doi={https://doi.org/10.1145/3534678.3539093}
}

@article{zhang2024situational,
  title={Situational-aware multi-graph convolutional recurrent network (SA-MGCRN) for travel demand forecasting during wildfires},
  author={Zhang, Xiaojian and Zhao, Xilei and Xu, Yiming and Nilsson, Daniel and Lovreglio, Ruggiero},
  journal={Transportation Research Part A: Policy and Practice},
  volume={190},
  pages={104242},
  year={2024},
  publisher={Elsevier},
  doi={https://doi.org/10.1016/j.tra.2024.104242}
}

@inproceedings{hu2024towards,
  title={Towards unifying diffusion models for probabilistic spatio-temporal graph learning},
  author={Hu, Junfeng and Liu, Xu and Fan, Zhencheng and Liang, Yuxuan and Zimmermann, Roger},
  booktitle={Proceedings of the 32nd ACM International Conference on Advances in Geographic Information Systems},
  pages={135--146},
  year={2024},
  doi={https://doi.org/10.1145/3678717.3691235}
}

@inproceedings{liu2023spatio,
  title={Spatio-temporal adaptive embedding makes vanilla transformer sota for traffic forecasting},
  author={Liu, Hangchen and Dong, Zheng and Jiang, Renhe and Deng, Jiewen and Deng, Jinliang and Chen, Quanjun and Song, Xuan},
  booktitle={Proceedings of the 32nd ACM international conference on information and knowledge management},
  pages={4125--4129},
  year={2023},
  doi={https://doi.org/10.1145/3583780.3615160}
}

@inproceedings{li2024m3,
  title={M3 LUC: Multi-modal Model for Urban Land-Use Classification},
  author={Li, Sibo and Zhang, Xin and Lin, Yuming and Li, Yong},
  booktitle={Proceedings of the 32nd ACM International Conference on Advances in Geographic Information Systems},
  pages={270--281},
  year={2024},
  doi={https://doi.org/10.1145/3678717.3691278}
}

@article{wang2023ssl4eo,
  title={SSL4EO-S12: A large-scale multimodal, multitemporal dataset for self-supervised learning in Earth observation},
  author={Wang, Yi and Braham, Nassim Ait Ali and Xiong, Zhitong and Liu, Chenying and Albrecht, Conrad M and Zhu, Xiao Xiang},
  journal={IEEE Geoscience and Remote Sensing Magazine},
  volume={11},
  number={3},
  pages={98--106},
  year={2023},
  publisher={IEEE},
  doi={10.1109/MGRS.2023.3281651}
}

@article{chen2024terra,
  title={Terra: A multimodal spatio-temporal dataset spanning the earth},
  author={Chen, Wei and Hao, Xixuan and Wu, Yuankai and Liang, Yuxuan},
  journal={Advances in Neural Information Processing Systems},
  volume={37},
  pages={66329--66356},
  year={2024},
}

@inproceedings{zhuang2024sauc,
  title={SAUC: Sparsity-Aware Uncertainty Calibration for Spatiotemporal Prediction with Graph Neural Networks},
  author={Zhuang, Dingyi and Bu, Yuheng and Wang, Guang and Wang, Shenhao and Zhao, Jinhua},
  booktitle={Proceedings of the 32nd ACM International Conference on Advances in Geographic Information Systems},
  pages={160--172},
  year={2024},
  doi={https://doi.org/10.1145/3678717.3691241}
}

@article{zhu2020beyond,
  title={Beyond homophily in graph neural networks: Current limitations and effective designs},
  author={Zhu, Jiong and Yan, Yujun and Zhao, Lingxiao and Heimann, Mark and Akoglu, Leman and Koutra, Danai},
  journal={Advances in neural information processing systems},
  volume={33},
  pages={7793--7804},
  year={2020},
}

@article{jiang2024self,
  title={Self-attention empowered graph convolutional network for structure learning and node embedding},
  author={Jiang, Mengying and Liu, Guizhong and Su, Yuanchao and Wu, Xinliang},
  journal={Pattern Recognition},
  volume={153},
  pages={110537},
  year={2024},
  publisher={Elsevier},
  doi={https://doi.org/10.1016/j.patcog.2024.110537}
}

@article{zhang2025linear,
  title={Linear attention based spatiotemporal multi graph GCN for traffic flow prediction},
  author={Zhang, Yanping and Xu, Wenjin and Ma, Benjiang and Zhang, Dan and Zeng, Fanli and Yao, Jiayu and Yang, Hongning and Du, Zhenzhen},
  journal={Scientific Reports},
  volume={15},
  number={1},
  pages={8249},
  year={2025},
  publisher={Nature Publishing Group UK London},
  doi={https://doi.org/10.1038/s41598-025-93179-y}
}

@article{huang2022unfolding,
  title={Unfolding community homophily in US metropolitans via human mobility},
  author={Huang, Xiao and Zhao, Yuhui and Wang, Siqin and Li, Xiao and Yang, Di and Feng, Yu and Xu, Yang and Zhu, Liao and Chen, Biyu},
  journal={Cities},
  volume={129},
  pages={103929},
  year={2022},
  publisher={Elsevier},
  doi={https://doi.org/10.1016/j.cities.2022.103929}
}

@article{yu2019exploring,
  title={Exploring the spatial variation of ridesourcing demand and its relationship to built environment and socioeconomic factors with the geographically weighted Poisson regression},
  author={Yu, Haitao and Peng, Zhong-Ren},
  journal={Journal of Transport Geography},
  volume={75},
  pages={147--163},
  year={2019},
  publisher={Elsevier},
  doi={https://doi.org/10.1016/j.jtrangeo.2019.01.004}
}

@article{lavieri2019modeling,
  title={Modeling individuals’ willingness to share trips with strangers in an autonomous vehicle future},
  author={Lavieri, Patr{\'\i}cia S and Bhat, Chandra R},
  journal={Transportation research part A: policy and practice},
  volume={124},
  pages={242--261},
  year={2019},
  publisher={Elsevier},
  doi={https://doi.org/10.1016/j.tra.2019.03.009}
}

@inproceedings{hajisafi2023learning,
  title={Learning dynamic graphs from all contextual information for accurate point-of-interest visit forecasting},
  author={Hajisafi, Arash and Lin, Haowen and Shaham, Sina and Hu, Haoji and Siampou, Maria Despoina and Chiang, Yao-Yi and Shahabi, Cyrus},
  booktitle={Proceedings of the 31st ACM International Conference on Advances in Geographic Information Systems},
  pages={1--12},
  year={2023},
  doi={https://doi.org/10.1145/3589132.362556}
}

@article{eren2020reviewBSP,
  title={A review on bike-sharing: The factors affecting bike-sharing demand},
  author={Eren, Ezgi and Uz, Volkan Emre},
  journal={Sustainable cities and society},
  volume={54},
  pages={101882},
  year={2020},
  publisher={Elsevier},
  doi={https://doi.org/10.1016/j.scs.2019.101882}
}

@article{petropoulos2022forecastingtheory,
  title={Forecasting: theory and practice},
  author={Petropoulos, Fotios and Apiletti, Daniele and Assimakopoulos, Vassilios and Babai, Mohamed Zied and Barrow, Devon K and Taieb, Souhaib Ben and Bergmeir, Christoph and Bessa, Ricardo J and Bijak, Jakub and Boylan, John E and others},
  journal={International Journal of forecasting},
  volume={38},
  number={3},
  pages={705--871},
  year={2022},
  publisher={Elsevier}
}

@article{zhao2019t,
  title={T-GCN: A temporal graph convolutional network for traffic prediction},
  author={Zhao, Ling and Song, Yujiao and Zhang, Chao and Liu, Yu and Wang, Pu and Lin, Tao and Deng, Min and Li, Haifeng},
  journal={IEEE transactions on intelligent transportation systems},
  volume={21},
  number={9},
  pages={3848--3858},
  year={2019},
  publisher={IEEE},
  doi={10.1109/TITS.2019.2935152}
}

@article{zou2024learning,
  title={Learning geospatial region embedding with heterogeneous graph},
  author={Zou, Xingchen and Huang, Jiani and Hao, Xixuan and Yang, Yuhao and Wen, Haomin and Yan, Yibo and Huang, Chao and Liang, Yuxuan},
  journal={arXiv preprint arXiv:2405.14135},
  year={2024}
}

@article{liu2020spatiotemporal,
  title={Spatiotemporal activity modeling via hierarchical cross-modal embedding},
  author={Liu, Yang and Ao, Xiang and Dong, Linfeng and Zhang, Chao and Wang, Jin and He, Qing},
  journal={IEEE Transactions on Knowledge and Data Engineering},
  volume={34},
  number={1},
  pages={462--474},
  year={2020},
  publisher={IEEE},
  doi={10.1109/TKDE.2020.2983892}
}

@misc{OpenStreetMap,
   author = {{OpenStreetMap contributors}},
   title = {{Planet dump retrieved from https://planet.osm.org }},
   howpublished = "\url{ https://www.openstreetmap.org }",
   year = {2017},
   lastchecked={June 5, 2025}
 }

@misc{openweather,
     author = {OpenWeatherMap},
     title={OpenWeatherMap API},
    howpublished = "\url{ https://openweathermap.org}",
   year={2025},
   lastchecked={June 5, 2025}
}

@techreport{homer2012national,
  title={The national land cover database},
  author={Homer, Collin G and Fry, Joyce A and Barnes, Christopher A},
  year={2012},
  institution={US Geological Survey}
}

@misc{uscb_acs,
    title={American Community Survey 5-Year Estimates},
    author={U.S. Census Bureau},
    year={2020},
    institution={US},
    howpublished = "\url{ https://data.census.gov/table/ACSST5Y2022}",
}

@misc{uscb_lehd,
    title={Longitudinal Employer-Household Dynamics},
    author={U.S. Census Bureau},
    year={2020},
    institution={US},
    howpublished = "\url{ https://lehd.ces.census.gov/data/}",
}

@inproceedings{lea2017temporal,
  title={Temporal convolutional networks for action segmentation and detection},
  author={Lea, Colin and Flynn, Michael D and Vidal, Rene and Reiter, Austin and Hager, Gregory D},
  booktitle={proceedings of the IEEE Conference on Computer Vision and Pattern Recognition},
  pages={156--165},
  year={2017}
}

@article{zhang2025towards,
  title={Towards Generalized Urban Computing: Pretraining a Spatial-Temporal Model for Diverse Urban Tasks},
  author={Zhang, Yingqian and Li, Chao and He, Shibo and Zhang, Xiangliang and Chen, Jiming},
  journal={IEEE Transactions on Mobile Computing},
  year={2025},
  publisher={IEEE},
  doi={10.1109/TMC.2025.3573373}
}

\end{document}